%% file: root.tex
\documentclass[twocolumn, switch]{article} 

\usepackage{color,array,amsthm,amsmath,amssymb,amsfonts}
\usepackage{graphicx}
\usepackage{booktabs}
\usepackage{pifont}
\usepackage{gensymb}
\usepackage{graphicx} 
\usepackage{nicematrix}
\usepackage{multirow}
\usepackage{makecell}
\usepackage{adjustbox} 

\input{tikz_preamble}

\usepackage{xspace}

\usepackage{svg}
\usepackage{bbding}
\usepackage{preprint}
\usepackage{xcolor}

\usepackage{siunitx}

\makeatletter
\DeclareRobustCommand\onedot{\futurelet\@let@token\@onedot}

\def\@onedot{\ifx\@let@token.\else.\null\fi\xspace}

\def\eg{\emph{e.g}\onedot} 
\def\ie{\emph{i.e}\onedot}

\makeatother

\usepackage{amsmath, amsthm, amssymb, amsfonts}

\usepackage[numbers,square]{natbib}
\usepackage[utf8]{inputenc}	
\usepackage[T1]{fontenc}	
\usepackage{xcolor}		
\usepackage[colorlinks = true,
            linkcolor = purple,
            urlcolor  = blue,
            citecolor = cyan,
            anchorcolor = black]{hyperref}	
\usepackage{booktabs} 		
\usepackage{nicefrac}		
\usepackage{microtype}		
\usepackage{lineno}		
\usepackage{float}			

\usepackage{balance}
\usepackage{lipsum}		

\usepackage{newfloat}
\DeclareFloatingEnvironment[name={Supplementary Figure}]{suppfigure}
\usepackage{sidecap}
\sidecaptionvpos{figure}{c}

\usepackage{titlesec}
\titlespacing\section{0pt}{12pt plus 3pt minus 3pt}{1pt plus 1pt minus 1pt}
\titlespacing\subsection{0pt}{10pt plus 3pt minus 3pt}{1pt plus 1pt minus 1pt}
\titlespacing\subsubsection{0pt}{8pt plus 3pt minus 3pt}{1pt plus 1pt minus 1pt}

\usepackage{tikz,xcolor,hyperref}

\usepackage{cleveref}

\definecolor{lime}{HTML}{A6CE39}
\DeclareRobustCommand{\orcidicon}{
	\begin{tikzpicture}
	\draw[lime, fill=lime] (0,0) 
	circle [radius=0.16] 
	node[white] {{\fontfamily{qag}\selectfont \tiny ID}};
	\draw[white, fill=white] (-0.0625,0.095) 
	circle [radius=0.007];
	\end{tikzpicture}
	\hspace{-2mm}
}
\foreach \x in {A, ..., Z}{\expandafter\xdef\csname orcid\x\endcsname{\noexpand\href{https://orcid.org/\csname orcidauthor\x\endcsname}
			{\noexpand\orcidicon}}
}

\title{CAD‑Free Learning of Spacecraft Pose Estimators via NeRF‑Based Augmentations}

\usepackage{authblk}
\author[1,2,4,5\thanks{\tt{antoine.legrand@uclouvain.be}}]{Antoine Legrand\orcidA{}}
\author[2,3,4]{Renaud Detry\orcidB{}}
\author[1]{Christophe De Vleeschouwer \orcidC{}}

\affil[1]{Department of Electrical Engineering (ELEN), ICTEAM, UCLouvain}
\affil[2]{Department of Electrical Engineering (ESAT), KU Leuven}
\affil[3]{Department of Mechanical Engineering (MECH), KU Leuven}
\affil[4]{Flanders Make@KU Leuven}
\affil[5]{Aerospacelab}

\begin{document}

\twocolumn[ 
  \begin{@twocolumnfalse} 
  
\maketitle

\begin{abstract}
Spacecraft pose estimation networks require tens of thousands of CAD‑rendered images to be trained. 
This reliance on synthetic CAD data (i) limits applicability to targets with reliable geometry prior, excluding uncooperative or poorly documented spacecraft, and (ii) causes poor generalization to real on‑orbit conditions due to unrealistic illumination and material appearance.

This paper introduces a NeRF‑based image augmentation method that enables the learning of spacecraft pose estimators from only a few tens to a few hundreds of images. 
The method learns a Neural Radiance Field of the target and generates a large, diverse dataset through geometrically‑consistent viewpoint and appearance augmentation. 
This augmented dataset enables the training of accurate target‑specific pose estimators without requiring a CAD model or large synthetic datasets.

Experiments show that our approach supports the training of accurate pose estimators from only 25 to 400 realistic images, even under severe illumination variations.
When applied on large CAD‑based synthetic datasets, the NeRF‑based augmentation also enhances out‑of‑domain generalization, yielding improved robustness to real on‑orbit conditions.

\end{abstract}
\vspace{0.35cm}

  \end{@twocolumnfalse} 
] 

\section{INTRODUCTION}
    O{\scshape n-orbit} Servicing (OOS)~\cite{henshaw2014darpa,pyrak2022mevs} and Active Debris Removal (ADR)~\cite{forshaw2016removedebris,biesbroek2021clearspace} missions require autonomous Rendezvous and Proximity Operations (RPO), during which a chaser spacecraft must estimate the six‑degree‑of‑freedom pose of an uncooperative target using only onboard sensors~\cite{ventura2016autonomous}.
    Among the available sensing modalities~\cite{opromolla2017review}, \eg, LiDAR~\cite{opromolla2017pose}, infrared sensors~\cite{shi2015uncooperative}, time‑of‑flight sensors~\cite{martinez2017pose}, event‑based cameras~\cite{jawaid2023towards}, and stereo cameras~\cite{pesce2017stereovision}, monocular cameras remain particularly attractive because of their low cost, mass, volume and power consumption~\cite{sharma2018pose,kisantal2020satellite,park2022speedplus}.
    
    Recent works have shown that learning‑based methods can achieve accurate monocular pose estimation~\cite{sharma2018pose,chen2019satellite,park2024robust,legrand2024domain}. 
    These approaches rely on training sets containing tens of thousands of labeled images that depict the target under a wide range of viewpoints and illumination conditions. 
    Since collecting such large real datasets before a mission is not possible, these training sets always consists in synthetic views rendered from a detailed Computer-Aided Design (CAD) model of the spacecraft.
    
    This training paradigm suffers from two fundamental limitations.
    First, the entire training process depends on the availability of an accurate CAD model. 
    Many OOS and ADR targets do not have publicly available geometry, and some are legacy or damaged spacecraft for which no reliable CAD exists. 
    Second,  synthetic images do not reproduce the lighting variability, specular reflections, and shadow dynamics encountered in orbit~\cite{park2021robotic,duzellier2022space}. 
    As a result, networks trained exclusively on synthetic data exhibit poor Out-Of-Domain (OOD) generalization capabilities, \ie, they often fail to generalize to real conditions.
    
    The present work proposes a different approach to building training sets for monocular pose estimation. 
    We introduce a NeRF‑based image augmentation method that requires only a few tens to a few hundreds of real images of the target. 
    As illustrated in \Cref{fig_taes_overview_sets}, a Neural Radiance Field (NeRF)~\cite{mildenhall2021nerf} $M_\Phi$ is learned from this small set $S_{\textnormal{Orig}}$ and used to synthesize a large set $S_{\textnormal{NeRF}}$ of geometrically consistent novel views of the target under diverse appearance configurations. 
    The pose estimation network $F_\Theta$ can then be learned on the augmented set $S_{\textnormal{Augm}}$ that combines the NeRF-based set with the original images.
    Because the network is trained on images that are diverse in terms of viewpoints and appearances, its OOD accuracy is significantly improved.

\input{tikz_augmentation_overview}

    This approach eliminates the dependency on a detailed CAD model.
    It allows learning‑based perception to be applied to spacecraft for which the geometry is unknown, incomplete, or proprietary. 
    This is highly relevant for ADR, inspection missions, and servicing scenarios that involve objects with limited priors.    
    Furthermore, when a CAD model is available, the NeRF-generated views can also complement the synthetic dataset, to improve robustness to unobserved lighting variations and to improve generalization to real on‑orbit imagery.
    
    This work builds on our preliminary works presented in a conferences and published as pre-prints~\cite{legrand2024leveraging,legrand2024domain_nerf}.
    We validate our NeRF-based augmentation method in two scenarios of interest for proximity operations, respectively assuming a small number of realistic images, and a large number of CAD-based synthetic views.
    Across both scenarios, NeRF‑based augmentation improves pose‑estimation accuracy and broadens the range of missions in which learning‑based RPO perception can be deployed.
    
    The rest of this paper is organized as follows.
    \Cref{sec_taes_rel_works} positions our works with respect to previous arts.
    \Cref{sec_taes_background} contains background material on the NeRF-based image synthesis.
    \Cref{sec_taes_method} describes our NeRF-based augmentation method.
    \Cref{sec_taes_datasets} presents the datasets used to validate the proposed method while \Cref{sec_taes_validation} evaluates the benefits brought by our method on domain generalization capabilities of a spacecraft pose estimation network under different data availability assumptions.
    Finally, \Cref{sec_taes_conclusion} concludes.

\section{RELATED WORKS}
\label{sec_taes_rel_works}

    Real on‑orbit images of a target spacecraft are difficult to acquire before the proximity phase of an OOS or ADR mission. 
    As a result, only a small set of real images might be available, which results in insufficient coverage of the possible viewpoint and appearance. 
    For this reason, most spacecraft pose estimation methods rely on a detailed CAD model to generate a large synthetic training dataset~\cite{sharma2018pose,chen2019satellite,park2024robust,legrand2024domain}. 
    This reliance on a CAD model limits applicability to missions where accurate geometry is available, which excludes many uncooperative, degraded, or poorly documented targets. 
    Furthermore, synthetic imagery has limited appearance diversity and does not reproduce the illumination and reflectance conditions encountered in orbit~\cite{park2023satellite}. 
    These two limitations, \ie, \textbf{data scarcity} and \textbf{limited appearance diversity}, are addressed by our NeRF-based solution. 
    They have also been studied in previous art, as detailed below.
    
\subsection{Mitigating Data Scarcity}

    Data scarcity primarily results in poor coverage of the viewpoint space, causing networks to extrapolate and overfit. 
    To address this, several geometric augmentation strategies have been proposed. 
    Two‑dimensional augmentations~\cite{wang2019perspective} such as cropping or scaling operate strictly in the image plane and provide limited viewpoint diversity. 
    Three‑dimensional approaches rely on multi‑camera setups~\cite{engilberge2023two} or generative models that synthesize new views, such as diffusion‑based~\cite{zhou2025generative} or adversarial methods~\cite{gong2021poseaug}. NeRF‑based techniques~\cite{feldmann2024nerfmentation,ruan2023towards,ge2022neural} also offer geometrically consistent novel views.  
    These \textit{methods improve viewpoint coverage}, but they focus exclusively on generating new views and \textit{do not address the limited appearance diversity} of the available images.

\begin{figure*}[t]
    \input{tikz_background}
    \caption{Image generation through a Neural Radiance Field (see \cref{sec_taes_background}) (reproduced from~\cite{legrand2026nerfreconstruction}). Each pixel value is predicted by casting a ray through the corresponding pixel in a camera of pose ($q$,$t$). $N$ points are sampled along the ray and fed in the neural field which predicts the point density and color. Using differential ray-tracing techniques to aggregate the sampled points, the pixel value can be computed.}
    \label{fig_back_nerf_rendering} 
\end{figure*}

\subsection{Increasing Appearance Diversity}

    The lack of appearance diversity in synthetic datasets causes pose estimation networks to rely on spurious visual cues and generalize poorly to real on‑orbit images. 
    Domain randomization~\cite{tobin2017domain} addresses this limitation by introducing variability during rendering. 
    CAD‑based pipelines~\cite{tremblay2018training,hinterstoisser2019annotation,zakharov2022photo} randomize texture, illumination, background, noise, or blur. 
    Other approaches introduce diversity directly in the image space, for example using style transfer~\cite{jackson2019style,yue2019domain}, random convolutions~\cite{xu2020robust}, randomized layers~\cite{lee2020network}, or Fourier perturbations~\cite{xu2021fourier}.
    Image‑space methods introduce appearance diversity without requiring a CAD model, but they remain restricted to two dimensional operations and cannot randomize the underlying three dimensional scene appearance.  
    
    Domain adaptation, which exploits information on the target domain such as unlabeled images to complement the source domain, has also been explored for spacecraft pose estimation~\cite{park2023satellite,wang2023bridging,perez2022spacecraft}. 
    However, these techniques require access to target‑domain data and incur computational costs that exceed the capabilities of current space‑grade hardware~\cite{cosmas2020fpga,leon2022fpga_asip}. 
    For operational missions, domain generalization through image augmentation therefore remains the only practically viable option.

\subsection{Summary}

    Existing approaches address viewpoint coverage or appearance diversity in isolation but do not overcome the fundamental limitations of current spacecraft pose estimation methods. 
    Many rely on CAD‑based rendering to compensate for data scarcity, yet these rendered datasets provide limited appearance variability. 
    Methods that introduce appearance diversity without a CAD model operate strictly within the two dimensional image grid, which constrains the range of variations they can generate. 
    As a result, none of these techniques can train a target‑specific pose estimator when only a small set of real images is available and no reliable CAD model exists.

\section{BACKGROUND}
\label{sec_taes_background}
    
    Neural Radiance Fields (NeRFs) represent a scene as a continuous volumetric function that predicts the density $\sigma$ and color $(r,g,b)$ at any 3D location.
    As illustrated in \Cref{fig_back_nerf_rendering}, an image $I$ of size $W \times H$ is rendered under a camera pose $(q,t)$ through differentiable ray-tracing.
    For each pixel $ij$, a ray is cast from the camera center with origin $c_{ij} = t$ and unit direction $d_{ij} = R(q)\, d^{\textnormal{P}}_{ij}$, where $d^{\textnormal{P}}_{ij}$ corresponds to the direction of a world-aligned reference camera.
    Along each ray, $N$ samples are queried, and their positions $(x,y,z)$ and viewing directions $(\theta,\phi)$ are passed to a neural field $\mathcal{F}$, which outputs $\sigma$ and $(r,g,b)$.
    The final pixel value is obtained by aggregating these predictions via differentiable ray-tracing.
    Despite architectural variations, most neural fields follow a similar structure in which spatial coordinates and viewing directions are encoded into feature vectors $F_{\textnormal{pos}}$ and $F_{\textnormal{dir}}$.
    An MLP processes $F_{\textnormal{pos}}$ to predict the density $\sigma$ and intermediate features $F_{\sigma}$.
    These features are concatenated with $F_{\textnormal{dir}}$ and fed into a second MLP to obtain the final color $(r,g,b)$.
        
    To address illumination variations in natural datasets, NeRF-in-the-wild~\cite{martin2021nerf} introduces learnable appearance embeddings, associating each training image with a vector $\mathbf{e}$ that is provided to the color MLP along with directional and density features.
    These embeddings model per-image appearance variations and improve robustness under varying lighting conditions.
    
    In this work, we adopt a NeRF based on the $K$-Planes architecture~\cite{fridovich2023k}, which combines efficient multi-plane interpolation with spherical-harmonic directional encoding while also supporting appearance embeddings.
    The resulting architecture, illustrated in \Cref{fig_back_nerf_rendering}, provides a good balance between computational efficiency and reconstruction fidelity.
    
    Because the image synthesis process is fully differentiable, the NeRF can be trained by minimizing a photometric loss between observed images $I_a$ and their synthesized ones under poses $(q_a,t_a)$ and appearance embeddings $\mathbf{e}_a$, \ie,
    \[
        \Phi^{*} = \operatorname*{argmin}_{\Phi} \mathcal{L}_{\textnormal{Photo}}\!\left(I_a,\: M_{\Phi}(q_a, t_a, e_a)\right).
    \]
    In practice, training operates on batches of rays corresponding to randomly sampled pixels $I_{aij}$, so that
    \[
        \Phi^{*} = \operatorname*{argmin}_{\Phi} 
        \mathcal{L}_{\textnormal{Photo}}\!\left(
            I_{aij},\:
            \mathcal{R}\!\big(\mathcal{F}(\mathcal{S}(c_{aij}, d_{aij}, e_a))\big)
        \right).
    \]
    Randomized pixel sampling exposes each batch to diverse viewpoints, promotes multi-view consistency, and accelerates convergence.

\section{METHOD}
\label{sec_taes_method}

\input{tikz_figure_method}

    This section presents our method for generating an augmented training set that provides broad viewpoint coverage and substantial appearance variability from an initially limited collection of images.
    By enriching the available data through this image generation pipeline, we enhance the OOD generalization capabilities of a pose estimation network trained on the resulting dataset.

    Our method relies on three successive stages.
    First, the input images are pre-processed to make them suitable for learning a neural scene representation.
    Second, two neural radiance fields are trained on the pre-processed data, as depicted in \Cref{fig_taes_method}.
    Finally, the first NeRF, $M_{\Phi}$, which captures the scene appearance, is in charge of generating foreground images from arbitrary viewpoints under diverse appearance conditions while the second NeRF, $M_{\Psi}$, which models the scene geometry, predicts the corresponding foreground masks.   
    When training the pose estimation network, the masks are used to insert a random background behind the foreground image in order to make the pose estimator robust to background distractors.

    The remainder of this section follows a three-stage structure.
    \Cref{sec_taes_meth_preprocess} describes the required pre-processing steps, and \Cref{sec_taes_meth_learning} details the learning strategy for training $M_{\Phi}$ and $M_{\Psi}$.
    Finally, \Cref{sec_taes_meth_app_rand} presents the image-synthesis procedure that preserves scene geometry while randomizing its appearance.

\subsection{Data Pre-processing}
\label{sec_taes_meth_preprocess}

    We first transform the available input images $\{I^{*}_{a}\}_{a=1}^{N_{\textnormal{orig}}}$ of width $W$ and height $H$ into a dataset that can be used for learning a NeRF. We start by estimating the relative pose $(\hat{q},\hat{t})_a$, and foreground mask $\hat{m}_a$ of each image $I^{*}_a$. 
    This can for example be done by using COLMAP~\cite{schonberger2016structure} or human annotations.
    The mask is then applied to remove the background from the original image in order to prevent the reconstruction of background content that is not relevant to the spacecraft.

    Using the camera calibration matrix, we cast rays through each pixel $ij$ of the image.
    Each ray is defined by the camera center $\hat{c}_a$ and the direction $\hat{d}_{aij}$ that corresponds to pixel $ij$.
    This produces a NeRF training set, in which each element consists of the ray origin $\hat{c}_a$, the ray direction $\hat{d}_{aij}$, the pixel value $I_{aij}$, and the binary mask value $m_{aij}$.

\subsection{Radiance Field Learning}
\label{sec_taes_meth_learning}

    We aim to learn a representation that can generate a novel view of the scene and the corresponding foreground mask for any given relative pose.
    In addition, the representation must allow the randomization of the scene appearance while preserving the underlying geometry.
    As discussed in \Cref{sec_taes_background}, Neural Radiance Fields (NeRFs) satisfy these requirements because they decouple scene geometry from appearance.

    Learning a NeRF from on-orbit imagery is challenging because illumination conditions vary strongly across views and because the relative pose labels contain uncertainty.
    To address these difficulties, we follow our previous work~\cite{legrand2026nerfreconstruction} and extend a NeRF with additional degrees of freedom.
    As introduced in~\cite{legrand2026nerfreconstruction}, per-image appearance embeddings allow the model to learn image-specific illumination conditions instead of averaging illumination across all views.
    Learnable pose correction terms enable the model to progressively refine the input pose labels and reduce the overall 3D uncertainty.

    As depicted in \Cref{fig_taes_method}, for each ray $(\hat{c}, \hat{d})$ in the NeRF training set, we refine the ray origin $\tilde{c}$ and the ray direction $\tilde{d}$ through a learnable pose correction term and we select the appearance embedding $\mathbf{e}$ associated with the image from which the ray is cast.
    We then sample $N_{\textnormal{P}}$ points along the ray.
    Each sampled point is characterized by its position $(x,y,z)$ and its two viewing angles $(\theta,\phi)$.
    For each point, the neural field predicts its density $\sigma$ and its color $y$.
    Differentiable ray-tracing techniques aggregate these quantities to predict the pixel value $I^{\textnormal{(photo)}}$.
    The photometric loss between the predicted and observed values is computed and back-propagated through the NeRF $M_{\Phi}$ to update its weights.
    After multiple iterations, $M_{\Phi}$ learns a representation of the scene geometry and appearance that enables the synthesis of novel views.

    Although $M_{\Phi}$ produces visually convincing images, preliminary experiments showed that the masks extracted from the predicted densities are noisy.
    Since these masks are later used to randomize the background or as supervision for pose estimator training, this level of noise is not acceptable.
    For this reason, we introduce a second NeRF, denoted $M_{\Psi}$, which focuses exclusively on the geometry.
    Its architecture is identical to that of $M_{\Phi}$, but the training objective differs.
    In addition to the photometric loss, we supervise the density using an auxiliary \emph{per-ray} loss $\mathcal{L}_{\sigma}$, defined based on the mask value $m(\mathbf{\hat{c}},\mathbf{\hat{d}})$ available for the ray of interest ($\mathbf{\hat{c}},\mathbf{\hat{d}}$),
    \begin{equation}
        \mathcal{L}_{\sigma} = \sum_{n=1}^{N_{\textnormal{P}}} \sigma_{n}^{2} \, (1 - m(\mathbf{\hat{c}},\mathbf{\hat{d}})).
    \end{equation}
    It enforces zero density for points that contribute only to background pixels.
    During training, $\mathcal{L}_{\sigma}$ is averaged over the mini-batch and combined with the photometric loss using unit weights, i.e., $\mathcal{L}_{\Psi}=\mathcal{L}_{\textnormal{photo}}+\mathcal{L}_{\sigma}$.
    This constraint prevents the network from hallucinating density outside the target envelope in order to improve the image reconstruction loss.
    As a consequence, the images synthesized by $M_{\Psi}$ have lower visual quality, but the masks extracted from them are significantly sharper.
    The pose‑correction terms are updated through $M_{\Phi}$; $M_{\Psi}$ does not contribute to these pose correction terms.

\begin{table*}[t]
    \setlength{\tabcolsep}{0pt}
    \renewcommand{\arraystretch}{1.15}
    \centering
    \caption{\small Overview of the 4 image sets used in our experiments.
    The 3 sets from SPEED+ are generated poses randomly sampled in $\mathrm{SE(3)}$ while SHIRT's ROE2 correspond to an approach trajectory.
    Images are either synthesized using the CAD model or captured in a HIL setup using a spacecraft mock-up.
    The HIL configuration enables two illumination domains: \textit{Lightbox}, reproducing diffuse Earth-albedo lighting, and \textit{Sunlamp}, emulating direct solar illumination. 
    To differentiate the SHIRT set, we add them the suffix "\_ROE2" to highlight its sequential nature while preserving the domain information (\textit{Lightbox\_ROE2}).}
    \label{tab_taes_dataset}
    \begin{NiceTabular}{|@{}>{\centering\arraybackslash}m{2.6cm}
        *{4}{|>{\centering\arraybackslash}m{3.05cm}}|}[
        cell-space-top-limit=0pt,
        cell-space-bottom-limit=0pt,
        code-before = \tikz{\node{};}
        ]
        \hline
        \centering Dataset 
            & \multicolumn{3}{c}{SPEED+~\cite{park2022speedplus}}
            & SHIRT~\cite{park2023adaptive}: ROE2\\
        \hline
        \centering Orig. Set Name
            & \textit{Synthetic}
            & \textit{Sunlamp}
            & \textit{Lightbox}
            & \textit{Lightbox}\\
        \hline
        \centering Pose distr.
            & \multicolumn{3}{c}{Random $\mathrm{SE(3)}$ Sampling}
            & Sampling over ROE \\
        \hline
        \centering Target
            & Simplified CAD
            & \multicolumn{3}{c}{Realistic mockup (HIL)} \\
        \hline
        \centering Illumination
            & OpenGL-based
            & Direct (Sun)
            & \multicolumn{2}{c}{Diffuse (Albedo)} \\
        \hline
        \centering \# images
            & 59,960
            & 2,791
            & 6,740
            & 2,371 \\
        \hline
        \adjustbox{valign=m}{Examples}
            & \raisebox{-0.145cm}{\smash{\includegraphics[width=\linewidth]{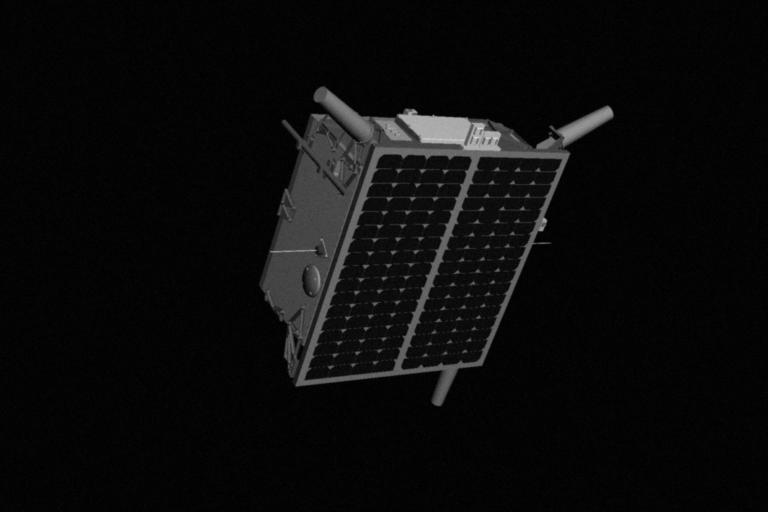}}}\rule{0pt}{1.875cm}
            & \raisebox{-0.145cm}{\smash{\includegraphics[width=\linewidth]{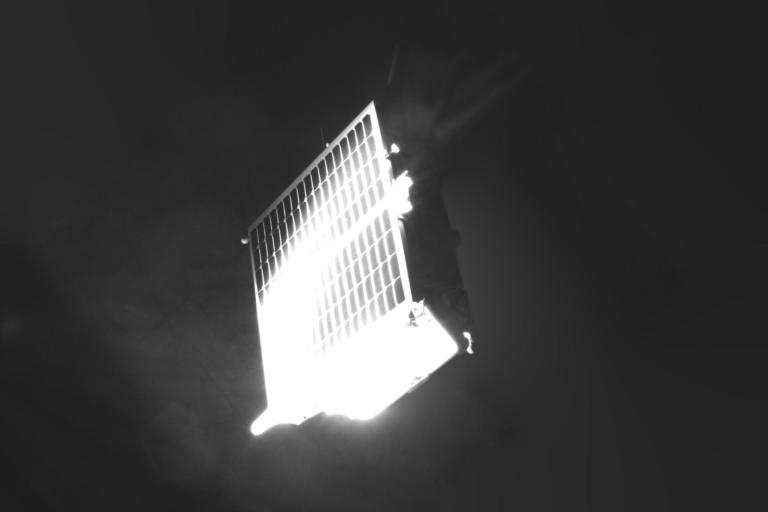}}}\rule{0pt}{1.875cm}
            & \raisebox{-0.145cm}{\smash{\includegraphics[width=\linewidth]{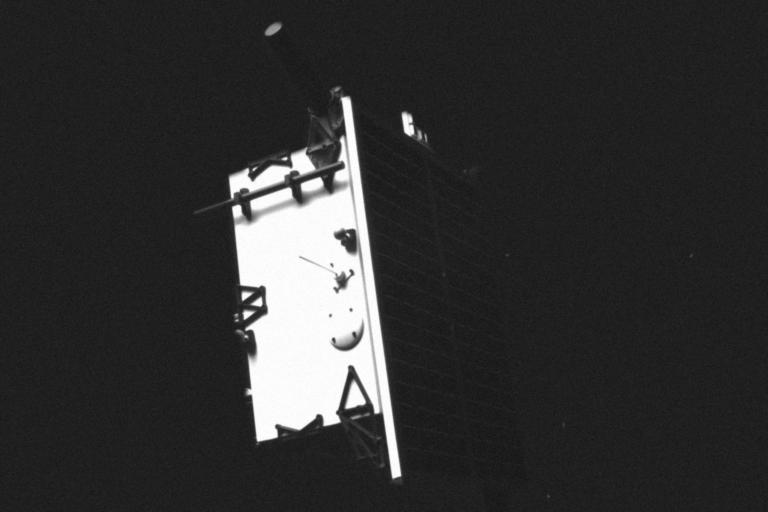}}}\rule{0pt}{1.875cm} 
            & \raisebox{-0.145cm}{\smash{\includegraphics[width=\linewidth]{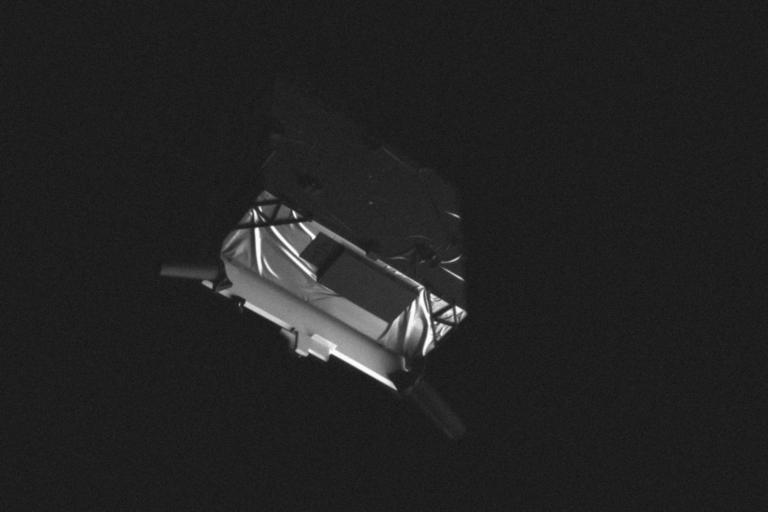}}}\rule{0pt}{1.875cm} \\
        \hline
    \end{NiceTabular}
\end{table*}

\subsection{Appearance-randomized Image Generation}
\label{sec_taes_meth_app_rand}

    To augment the training set with both viewpoint diversity and appearance variability, we begin by randomly sampling $N_{\textnormal{lab}}$ pose labels in $\mathrm{SE}(3)$.
    For each sampled pose $(q,t)$, we render the foreground mask $m$ with the density-supervised NeRF $M_{\Psi}$ by thresholding the accumulated opacity at $\tau$, and we synthesize $N_{\textnormal{cfg}}$ foreground images for that viewpoint with the appearance-supervised NeRF $M_{\Phi}$, reusing its density and density features while randomizing appearance.
    
    For this purpose, we use the sampler and the density field of $M_{\Phi}$ to compute the density $\sigma$ together with the direction and density features $F_{\textnormal{dir}}$ and $F_{\sigma}$.
    Since the density field defines the projection of the scene geometry into the image, all remaining steps of the image generation process can be randomized without altering the target shape.
    We rely on two mechanisms to randomize the appearance: illumination randomization through the appearance embedding and color randomization through the color network.

    \paragraph{Illumination Randomization}
    During training, the network learns a set of appearance embeddings $\mathcal{E}=\{\mathbf{e}_1,\dots,\mathbf{e}_N\}\subset\mathbb{R}^d$.
    At inference time, we randomize the appearance by sampling a new embedding $\mathbf{e}^{(k)}$ using one of four strategies.
    The first strategy selects an embedding uniformly at random, meaning that we draw $i\sim\mathrm{Unif}\{1,\dots,N\}$ and set $\mathbf{e}^{(k)}=\mathbf{e}_i$.
    The second strategy interpolates between two embeddings by sampling $i\neq j$ and a coefficient $\alpha\in[0,1]$ and forming
    \begin{equation}
        \mathbf{e}^{(k)} = \mathbf{e}_i + \alpha(\mathbf{e}_j - \mathbf{e}_i).
    \end{equation}
    The third strategy extrapolates between two embeddings using the same expression but with a coefficient $\alpha$ outside the interval $[0,1]$.
    The final strategy samples $\mathbf{e}^{(k)}$ from a Gaussian model fitted to $\mathcal{E}$, treating the learned embeddings as realizations of a multivariate normal distribution with empirical mean $\boldsymbol{\mu}$ and covariance $\boldsymbol{\Sigma}_{\textnormal{emb}}$, and drawing $\mathbf{e}^{(k)} \sim \mathcal{N}(\boldsymbol{\mu}, \boldsymbol{\Sigma}_{\textnormal{emb}})$.

    \paragraph{Color Randomization}
    A second approach for appearance randomization consists in injecting Gaussian noise into the parameters of the color MLP.
    Instead of using the deterministic color network $f_{\phi^{\textnormal{(col)}}}^{\textnormal{(col)}}$, we perturb its parameters with a noise vector $\varepsilon^{(k)} \sim \mathcal{N}(\mathbf{0},\boldsymbol{\Sigma}_{\Phi})$ and evaluate the colors using the perturbed network $f^{\textnormal{(col)}}_{\phi^{\textnormal{(col)}} + \varepsilon^{(k)}}$.
    The color of each sampled point is then computed using this noise-injected parameter vector.

    The point-wise colors $\{y_n^{(k)}\}$ predicted by the randomized color network are combined with the preserved densities $\{\sigma_n\}$ to compute the final pixel value $I^{(k)}$.
    By applying different randomization configurations $(\mathbf{e}^{(k)}, \varepsilon^{(k)})$, we generate $N_{\textnormal{cfg}}$ images of the target spacecraft that all correspond to the same pose but exhibit different appearances.
    
\section{DATASETS}
\label{sec_taes_datasets}

    Our experiments are conducted on 4 image sets from SPEED+~\cite{park2022speedplus} (\textit{Synthetic}, \textit{Lightbox}, \textit{Sunlamp}) and SHIRT~\cite{park2023adaptive} (\textit{Lightbox\_ROE2}).
    They are briefly summarized in \Cref{tab_taes_dataset}.
    Both datasets have been generated using the same rendering tool and acquisition setup.
    They contain 1200$\times$1920 grayscale images of \textsc{Tango} from the PRISMA mission~\cite{gill2007autonomous}, along with the corresponding pose labels.
    
    They contain both synthetic and Hardware-In-the-Loop (HIL) images captured at the TRON facility at Stanford~\cite{park2021robotic}. 
    The HIL acquisition setup emulates the illumination conditions encountered in orbit on a spacecraft mockup. Two HIL domains are generated. \textit{Lightbox} mimics a diffuse illumination resulting from from the Earth albedo while \textit{Sunlamp} reproduces the Sun direct illumination on the spacecraft's surfaces.    
    \textit{Synthetic} images are rendered using an OpenGL-based renderer that emulates the camera setup used in the HIL setup. Due to the lack of details within the CAD model and the tool rasterization approach, synthetic images do not capture the appearance conditions encountered in the HIL images.

    While image sets of SPEED+ correspond to \textit{Synthetic}, \textit{Lightbox} and \textit{Sunlamp} domains, with pose labels randomly sampled in $\mathrm{SE(3)}$, the \textit{Lightbox\_ROE2} set of SHIRT corresponds to a slow approach of a chaser towards a tumbling spacecraft under diffuse illumination. 

    Since \textit{Lightbox\_ROE2} is close to a realistic in-orbit acquisition scenario through its trajectory-based pose distribution and illumination conditions, we exploit its images as the original train set $S_{\textnormal{Orig}}$. As the HIL sets of SPEED+ (\textit{Lightbox} and \textit{Sunlamp}) depict realistic illumination while being independent from \textit{Lightbox\_ROE2}, they are used to evaluate the domain generalization abilities of a network trained using our augmentation method.
    We also conduct similar experiments on models trained using the \textit{Sunlamp} and \textit{Synthetic} sets. 
    The former enables the evaluation of the method's ability to deal with more challenging illumination conditions.
    The latter evaluates the method's ability in improving the test-time accuracy of a pose estimator trained on a large set of synthetic images generated using the target CAD model.

\section{EXPERIMENTAL VALIDATION}
\label{sec_taes_validation}

    This section describes the experiments conducted to validate our proposed augmentation method.
    First, \Cref{sec_taes_val_setup} presents our experimental setup. 
    Then, \Cref{sec_taes_val_main} demonstrates how our method enables the training of a pose estimator on a dataset with severe limitations regarding the number of available viewpoints as well as the image diversity.
    \Cref{sec_taes_val_nimages} and \Cref{sec_taes_val_sunlamp} then evaluates our method's capabilities when applied on datasets that are even more limited in terms of number of viewpoints and quality of the images, respectively.
    To isolate the performance's improvement originating from the increase in number of viewpoints, we then show in \Cref{sec_taes_val_synthetic} how our method improves the Out-Of-Domain generalization capabilities of a pose estimation network when an arbitrary large number of viewpoints are available, \eg, because a CAD model is available. 
    Finally, \Cref{sec_taes_val_abl_density_sup} performs an ablation study on the use of the second NeRF dedicated to the synthesis of segmentation masks.

    \subsection{Experimental Setup}
    \label{sec_taes_val_setup}

    Our method enables the training of pose estimation network $F_\Theta$ from a set of images that is too scarce and lacks of appearance diversity.   
    Unless otherwise mentioned, this original set $S_\textnormal{Orig}$ consists in $N_{\textnormal{Orig}}$=400 images of 1920$\times$1200 pixels randomly sampled in \textit{Lightbox\_ROE2}. 
    As explained in \Cref{sec_taes_datasets}, \textit{Lightbox\_ROE2} is representative of the data that could be acquired in a real rendezvous operation because it was acquired with a setup that emulates the slow approach of a chaser spacecraft towards a tumbling target under illumination conditions that reproduce the ones encountered in orbit.
    As for real data, the pose labels of \textit{Lightbox\_ROE2} are noisy because of the testbed calibration uncertainty.

    We train two NeRFs, $M_{\Phi}$ and $M_{\Psi}$, focusing respectively on the learning the object appearance and on the scene density.
    Both NeRFs follow the $K$-Planes implementation~\cite{fridovich2023k}.
    Once trained, they are used to generate the augmented set $S_{\textnormal{NeRF}}$ by sampling 48,000 pose labels in $\mathrm{SE(3)}$. 
    In practice, these labels are taken from the train set of the \textit{Synthetic} set to provide a ground for comparison.
    For each pose label, a segmentation mask and $N_{\textnormal{cfg}}=64$ images corresponding to that pose under different appearance-randomization configurations (24 illumination augmentations ($\mathbf{e}^{(k)}$) and 40 color augmentations ($\epsilon^{(k)}$)) are generated.
    Training both NeRFs takes 50 minutes on a NVIDIA L40s while the rendering of each mask takes 1.5 second and the rendering of the $N_{\textnormal{cfg}}=64$ images takes 3.8 seconds.
    Both images and masks are generated at a 768$\times$512 resolution which corresponds to the input resolution of the pose estimation network $F_{\Theta}$.   

    \begin{figure}[t]
        \centering
        \resizebox{\linewidth}{!}{\input{tikz_spnv2.tex}}
        \caption{Architectural overview of SPNv2~\cite{park2024robust}. 
        The network extracts multi‑scale feature maps using an EfficientNet backbone and a BiFPN. They are processed through three heads: (i) a pose‑regression head, (ii) a keypoint head whose predictions are used by a PnP solver to compute the pose, and (iii) a segmentation head providing auxiliary supervision. The model outputs a hybrid pose combining the regressed translation $t^{\text{reg}}$ with the keypoint‑based orientation $q^{\text{heat}}$.}
        \label{fig_spe_spnv2}
    \end{figure}

    The pose estimation network $F_\Theta$ used in our experiments is SPNv2~\cite{park2024robust}, a state-of-the-art spacecraft pose estimation network designed to achieve good out-of-domain generalization capabilities, and is depicted in \Cref{fig_spe_spnv2}.
    SPNv2 is  built around an EfficientNet~\cite{tan2019efficientnet} backbone and a BiFPN~\cite{tan2020efficientdet} for multi‑scale feature fusion, followed by three heads: an EfficientPose~\cite{bukschat2020efficientpose} regression head, a keypoint‑prediction head refined through a PnP solver~\cite{lepetit2009ep}, and a segmentation head used for auxiliary supervision. 
    At inference, the model outputs a hybrid pose in which the translation comes from the regression head while the orientation is taken from the PnP‑based branch.
    Designed for robustness under domain shift through multi‑task learning and augmentations, SPNv2 is used as a baseline to assess the impact of our augmentation method on out‑of‑domain generalization.

    In our experiments, SPNv2 is trained for 20 epochs where each epoch consists of the $N_{\textnormal{Orig}}$ images from the $S_{\textnormal{Orig}}$ set combined with the 48,000 new pose labels used to generate $S_{\textnormal{NeRF}}$. 
    For each pose label, we randomly sample one of the $N_{cfg}$ images generated for that pose label so that the network is exposed to different appearance-randomization configurations.
    We add the Earth in the background of half of the 48,000 images using the foreground segmentation masks predicted by $M_\Psi$. 
    These masks are also used a supervision signals for training SPNv2, as in the original paper~\cite{park2024robust}.

    Once trained, the pose estimation networks are evaluated on the \textit{Lightbox} and \textit{Sunlamp} datasets using as metrics the average translation error $E^{\textnormal{T}}$, angular error $E^{\textnormal{R}}$ and SPEED score $S^{P}$ which aggregates both translation and angular errors. For further details on these metrics, the reader is referred to~\cite{park2022speedplus}.

\subsection{CAD-free learning of a pose estimator}
\label{sec_taes_val_main}

    This section evaluates our method's ability of generating a sufficiently large and diverse dataset so as to enable the training of a robust spacecraft pose estimation network from a reduced set of images taken in orbit, thereby alleviating the CAD model requirement encountered by such models.
    Such a set suffers from limitations regarding both its viewpoint coverage and   appearance diversity.
    For this purpose, we conduct our experiments on a subset of \textit{Lightbox\_ROE2} containing 400 images taken under realistic illumination conditions and viewpoints that correspond to a slow approach towards a tumbling target spacecraft.

    We first trained the pose estimator on these 400 images using the standard training procedure of SPNv2. 
    Unsurprisingly, the so-trained network performs extremely poorly, with pose scores in the order of 2.0, which roughly corresponds to 80\degree, on all test sets.
    This indicates that the network completely overfitted the 400 images.     
    \Cref{tab_taes_expes_main} indicates the test-time performance of SPNv2 trained using different configurations of our NeRF-based augmentation method designed to prevent that overfitting.
    
    First, by generating 48,000 images using only the novel view synthesis ability of a NeRF trained on these 400 images, we can learn a pose estimation network properly. 
    It achieves a fair pose score of 0.168 on \textit{Lightbox}, but is significantly worse on \textit{Sunlamp} as the pose score rises to 0.529.
    This is coherent because the illumination conditions within \textit{Lightbox\_ROE2} are close to the ones encountered in \textit{Lightbox} but different from the challenging illumination conditions encountered in \textit{Sunlamp}.

\begin{table}[t]
\small
\centering
\setlength{\tabcolsep}{4pt}
\renewcommand{\arraystretch}{1.0}
\caption{OOD performance comparison of SPNv2 trained on the \textit{Lightbox\_ROE2} set under four augmentation strategies: no appearance, illumination‑only, color‑only, and combined illumination–color, augmentations. Applying both illumination and color augmentations yields the strongest OOD generalization performance.}
\label{tab_taes_expes_main}
\begin{tabular}{c|c|rrr|rrr}
    \toprule
    Ill. & Color & \multicolumn{3}{c|}{\textit{Lightbox}} & \multicolumn{3}{c}{\textit{Sunlamp}} \\
    Augm. & Augm. & $S^{\textnormal{P}} [/]  $ & $E^{\textnormal{R}}[\degree] $ & $E^{\textnormal{T}}[cm] $ & $S^{\textnormal{P}} [/]  $ & $E^{\textnormal{R}}[\degree] $ & $E^{\textnormal{T}}[cm]$ \\
    \midrule
    \ding{55} & \ding{55} & 0.168 & 7.55 & 23.43 & 0.529 & 25.62 & 57.12 \\
    \midrule
    \ding{51}& \ding{55} & 0.141 & 6.53 & 17.73 & 0.396 & 19.53 & 36.58 \\
    \ding{55} & \ding{51} & 0.144 & 6.31 & 21.37 & 0.276 & 12.90 & 35.61 \\
    \midrule
    \ding{51} & \ding{51} & \textbf{0.109} & \textbf{4.80} & \textbf{15.37} & \textbf{0.191} & \textbf{9.01} & \textbf{23.33} \\
    \bottomrule
\end{tabular}
\end{table}

    Then, \Cref{tab_taes_expes_main} also presents the performance metrics achieved by a pose estimation network trained using our appearance-diversification strategies. Illumination augmentations through randomized appearance embeddings lead to an improvement of the generalization capabilities on both sets. 
    On \textit{Lightbox} and \textit{Sunlamp}, the pose scores are of 0.141 and 0.396, respectively.
    These improvements are even better using appearance augmentations enabled by the randomization of the color MLP. 
    Using only these color augmentations, we achieve pose scores of 0.144 and 0.276, respectively.

    Finally, \Cref{tab_taes_expes_main} highlights that combining both forms of appearance augmentations leads to the best generalization capabilities.
    On \textit{Lightbox}, we achieve a pose score of 0.109 while on \textit{Sunlamp}, the pose score reaches 0.191.
    This indicates that the proposed geometry-preserving appearance-randomization strategy reduces by 35\% and 57\% the pose scores on \textit{Lightbox} and \textit{Sunlamp}, respectively, compared to performing only NeRF-based viewpoint augmentation.

    \subsection{On the impact of smaller train sets}
    \label{sec_taes_val_nimages}

    In the previous section, we evaluated the ability of our augmentation method to enable the learning of a pose estimation network from images acquired during a rendezvous approach.
    It assumed the access to a reasonably large set of 400 images taken in-orbit and depicting the target spacecraft.
    In practice, acquiring these 400 images of the target might be complicated.
    Hence, in this section, we assess the evolution of these capability when the number of available images is reduced. 

    \Cref{tab_taes_expes_nframes} compares the performance achieved by a pose estimation network trained using the same viewpoint and appearance strategy on 5 datasets of decreasing sizes. 
    As the number of available images decreases the pose score is progressively degraded.
    This is perfectly coherent as the level of information in the train set is reduced and the quality of the learned representation is therefore affected. 
    This decrease is however smooth and even at 100 images, the network generalization abilities remain attractive. 
    When decreasing the number of images from 400 to 100 images, the pose scores evolve from 0.109 on \textit{Lightbox} and 0.191 on \textit{Sunlamp} sets to 0.205 and 0.241, respectively.
    Reducing by a factor 4 the number of images that needs to be acquired therefore only costs an increase of the pose score of 88\% and 26\%, respectively on those sets. 
    
    If we further decrease the number of images, the performance drop but even when only 25 images are available, the image-to-pose mapping can still be learned, even if its accuracy on both domains is limited. 
    Although the number of available images is divided by a factor 16, the pose scores evolve from 0.109 and 0.191 to 0.403 and 0.282, respectively on both sets.
    While this corresponds to an increase of 270\% on \textit{Lightbox}, this increase is limited to 48\% on \textit{Sunlamp}. 
    These experiments therefore demonstrate that our method is able to cope with the really low data regimes that could be required in the context of proximity operations with a poorly defined target spacecraft.

\begin{table}[t]
\setlength{\tabcolsep}{4pt}
\renewcommand{\arraystretch}{1.0}
\small
\centering
\caption{OOD performance comparison of SPNv2 trained on five datasets of decreasing sizes augmented through the same viewpoint and appearance augmentations. The performance is smoothly reduced when the number of available images decreases.}
\label{tab_taes_expes_nframes}
\begin{tabular}{r|rrr|rrr}
    \toprule
    \multirow{2}{*}{$N_{\textnormal{Orig}}$} & \multicolumn{3}{c|}{\textit{Lightbox}} & \multicolumn{3}{c}{\textit{Sunlamp}} \\
    & $S^{\textnormal{P}} [/]  $ & $E^{\textnormal{R}}[\degree] $ & $E^{\textnormal{T}}[cm] $ & $S^{\textnormal{P}} [/]  $ & $E^{\textnormal{R}}[\degree] $ & $E^{\textnormal{T}}[cm]$ \\
    \midrule
    400 & \textbf{0.109} & \textbf{4.80} & \textbf{15.37} & \textbf{0.191} & \textbf{9.01} & 23.33 \\
    200 & 0.173 & 7.75 & 23.70 & 0.297 & 13.97 & 38.03 \\
    100 & 0.205 & 9.66 & 22.39 & 0.241 & 11.68 & 24.98 \\50 & 0.241 & 11.07 & 30.13 & 0.394 & 19.00 & 44.76 \\
    25 & 0.403 & 19.25 & 42.52 & 0.282 & 14.24 & \textbf{21.22} \\    
    \bottomrule
\end{tabular}
\end{table}

    \subsection{On the impact of adverse illumination conditions}
    \label{sec_taes_val_sunlamp}

    In the previous section, we examined our method's behavior when the training set is even smaller. 
    Beyond quantity, image quality is also of a tremendous importance.
    Indeed, we previously conducted our experiments on image sets acquired using a testbed emulating on-orbit illumination under diffuse conditions. 
    In practice, on-orbit images may be significantly affected by the Sun direct exposure and exhibit strong contrasts. 
    In this section, we therefore evaluate our method's resilience to harsh illumination conditions within the available training set.
    
    For this purpose, we carried out an experiment using 400 images randomly sampled in \textit{Sunlamp} to train the pose estimation network.
    The so-trained pose estimation network achieves good performances given the poor visual quality of the available images. 
    On \textit{Lightbox}, the pose score is equal to 0.291 while it is equal to 0.096 when computed on the \textit{Sunlamp} images that are not used in the training set. These images therefore share similar illumination conditions as the training set but are not seen by the pose estimation network at training. 

    This experiment demonstrates that the proposed augmentation method is able to generate an augmented set that is sufficiently rich in terms of viewpoints and appearances so as to enable the training of a robust pose estimation network even on datasets made of images of a poor visual quality. 
    

    \subsection{Augmentation of synthetic train sets}
    \label{sec_taes_val_synthetic}

    Our augmentation method addresses the two limitations that prevent the training of a pose estimation network from a small set of real images depicting the target spacecraft and impose the reliance on CAD model to generate a relevant training set.
    It enables both the generation of novel viewpoints as well the randomization of the target appearance through an implicit 3D model learned from these in-orbit images.
    In the previous sections, we demonstrated the coupled impact of both augmentation types.
    In this section, we aim at decoupling the impact of both augmentations. 
    We therefore carry our experiments on a training set made of 48,000 images from the \textit{Synthetic} set.
    With this large set, we do not need to generate any novel viewpoint, and therefore only render the current ones under different appearances so as to only evaluate the impact of the appearance augmentation.
    
\begin{table}[t]
\setlength{\tabcolsep}{4pt}
\renewcommand{\arraystretch}{1.0}
\small
\centering
\caption{OOD performance comparison of SPNv2 trained on 48,000 \textit{Synthetic} images with and without using our NeRF-based appearance-randomization strategy. Applying our augmentation strategy improves the out-of-domain generalization capabilities.}
\label{tab_taes_expes_synth}
\begin{tabular}{c|rrr|rrr}
    \toprule
    Augmentation & \multicolumn{3}{c|}{\textit{Lightbox}} & \multicolumn{3}{c}{\textit{Sunlamp}} \\

    Strategy & $S^{\textnormal{P}} [/]  $ & $E^{\textnormal{R}}[\degree] $ & $E^{\textnormal{T}}[cm] $ & $S^{\textnormal{P}} [/]  $ & $E^{\textnormal{R}}[\degree] $ & $E^{\textnormal{T}}[cm]$ \\
    \midrule
    No & 0.193 & 8.90 & \textbf{22.98} & 0.295 & 14.56 & 25.73 \\
    Color \& Ill. & \textbf{0.155} & \textbf{6.75} & 24.36 & \textbf{0.183} & \textbf{8.96} & \textbf{17.28} \\
    \bottomrule
\end{tabular}
\end{table}

    \Cref{tab_taes_expes_synth} compares the out-of-domain generalization capabilities of a pose estimation network trained either on the \textit{Synthetic} set or using that set augmented with our appearance-randomization method.
    On \textit{Lightbox}, our method significantly reduces the pose score by 19.7\%. 
    On \textit{Sunlamp}, this improvement is even more pronounced with a pose score improved by 38\%. 
    This highlights that the potential of our method is not only limited to enabling the training of pose estimators without relying on a CAD model.
    With this experiment, we showed that it can also improve the out-of-domain generalization capabilities of a pose estimator trained on a CAD-based dataset, even if that pose estimator already exhibit strong out-of-domain generalization capabilities through the use of a wide collection of image augmentation techniques.

    \subsection{Ablation on density supervision}
    \label{sec_taes_val_abl_density_sup}

    This section presents an ablation study on the use of the second NeRF $M_\Phi$ dedicated to the representation of the target density and the synthesis of the foreground masks $m$ used to (i) supervise the auxiliary segmentation task and (ii) enable the randomization of the image background so as to improve the network resilience to the background content, \eg, the Earth.

    \Cref{tab_taes_expes_ablation} compares performance metrics achieved by the pose estimation network when trained with our augmentation strategy using different mask synthesis strategies.
    It highlights the metrics achieved by the estimation network when trained using (i) the masks generated by $M_\Psi$, (ii) the masks generated by $M_\Phi$, or, (iii) no mask at all.
    Compared to using clean segmentation masks produced by $M_\Psi$, using those predicted by $M_\Phi$ degrades the pose scores as they evolve from 0.109 and 0.191 to 0.189 and 0.253, respectively on both sets. 
    Using no masks at all, \ie, neither for auxiliary segmentation nor for background augmentation, the pose scores on both sets are significantly degraded as they reach 0.424 and 0.387, respectively.
    This highlights the importance of segmentation masks when learning a pose estimation network and that their quality is so critical that we need a second NeRF $M_\Psi$ dedicated to their generation.

    \begin{table}[t]
    \setlength{\tabcolsep}{4pt}
    \renewcommand{\arraystretch}{1.0}
    \small
    \centering
    \caption{OOD performance comparison of SPNv2 trained using masks generated by either $M_\Psi$, the NeRF dedicate to geometry modeling, $M_\Phi$, the NeRF dedicated to appearance modeling, or using no mask at all. Using masks generated by $M_\Psi$ leads to the best performance.}
    \label{tab_taes_expes_ablation}
    \begin{tabular}{c|rrr|rrr}
        \toprule
        Masks & \multicolumn{3}{c|}{\textit{Lightbox}} & \multicolumn{3}{c}{\textit{Sunlamp}} \\
        Origin & $S^{\textnormal{P}} [/]  $ & $E^{\textnormal{R}}[\degree] $ & $E^{\textnormal{T}}[cm] $ & $S^{\textnormal{P}} [/]  $ & $E^{\textnormal{R}}[\degree] $ & $E^{\textnormal{T}}[cm]$ \\
        \midrule
        $M_{\Psi}$ & \textbf{0.109} & \textbf{4.80} & \textbf{15.37} & \textbf{0.191} & \textbf{9.01} & \textbf{23.33} \\
        $M_{\Phi}$ & 0.189 & 8.20 & 29.12 & 0.253 & 11.88 & 32.27 \\
        No masks & 0.424 & 19.25 & 59.04 & 0.387 & 18.15 & 52.23 \\  
        \bottomrule
    \end{tabular}
    \end{table}

\section{CONCLUSION}
\label{sec_taes_conclusion}

    This paper introduced a new approach to spacecraft pose estimation that removes the need for a CAD model and enables target-specific networks to be learned from only a small set of real or real-like images. 
    The core idea is to exploit a Neural Radiance Field learned from these images as a three dimensional scene representation in which geometry and appearance are naturally decoupled. 
    This representation allows our method to generate novel viewpoints and to randomize appearance directly in the scene without altering the underlying geometry.
    
    By combining viewpoint augmentation with appearance randomization, the proposed approach produces an augmented training set that can be used to train spacecraft pose estimation networks for targets whose geometry is unknown, undocumented, or inaccessible. 
    The same mechanism also improves Out-of-Domain generalization when used to augment large CAD-based datasets, since it introduces appearance variability that cannot be produced by conventional rendering or two dimensional image transformations.
    
    We demonstrated that robust pose estimation networks can be trained from a few tens to a few hundreds of images, including images captured under challenging illumination conditions. 
    Finally, a further analysis highlighted the importance of accurate segmentation masks produced by a second NeRF model dedicated to modeling the scene's geometry.

\section*{Acknowledgements}
The research was funded by Aerospacelab and the Walloon Region through the Win4Doc program. Christophe De Vleeschouwer is a Research Director of the Fonds de la Recherche Scientifique - FNRS. Computational resources have been provided by the Consortium des Équipements de Calcul Intensif (CÉCI), funded by the Fonds de la Recherche Scientifique de Belgique (F.R.S.-FNRS) under Grant No. 2.5020.11 and by the Walloon Region.

The authors used MS Copilot during the preparation of this manuscript to assist with figure generation from textual descriptions (Figures 1$\rightarrow$4), suggest alternative phrasings, and perform minor language corrections. All scientific content, methodology, and conclusions were developed and validated by the authors, who take full responsibility for the work.

\normalsize
\bibliography{root}


\end{document}

%% file: tikz_preamble.tex
\usepackage[dvipsnames]{xcolor} 
\usepackage{tikz}
\usetikzlibrary{positioning, calc, fit, arrows.meta, shapes, matrix,backgrounds}

\newlength{\setmargin}
\newlength{\cellw}
\newlength{\cellh}
\newlength{\gridsep}
\newlength{\netdrop}
\newlength{\posenetdrop}

\newcommand{\imgcell}[1]{\includegraphics[width=\cellw,height=\cellh]{#1}}

\tikzset{
  block/.style={draw, rounded corners=2pt, thick, fill=blue!5},
  net/.style={block, fill=orange!10, minimum height=8mm, align=center},
  arrow/.style={-Latex, thick},
  setlabel/.style={font=\small\bfseries, inner sep=2pt, anchor=north west},
  smalltext/.style={font=\footnotesize},
}

\tikzset{
  dblarrow/.style={
    double=white,          
    double distance=1.2pt, 
    line width=0.6pt,      
    draw=black,              
    -{Latex[length=3mm,width=1.6mm]}, 
  },
  dblarrow tight/.style={dblarrow, double distance=0.8pt},
  dblarrow wide/.style={dblarrow, double distance=1.8pt},
  dblarrow gray/.style={dblarrow, draw=black!25},
}

%% file: tikz_augmentation_overview.tex
\begin{figure}[t]
\centering
\begin{tikzpicture}[font=\footnotesize]

\pgfdeclarelayer{infront}
\pgfsetlayers{background,main,infront}

\begin{pgfonlayer}{infront}
    \matrix (grid1) [matrix of nodes,
      nodes={inner sep=0pt, anchor=center},
      column sep=\gridsep, row sep=\gridsep,
      anchor=north west
    ] at (0,0) {
      \node{\imgcell{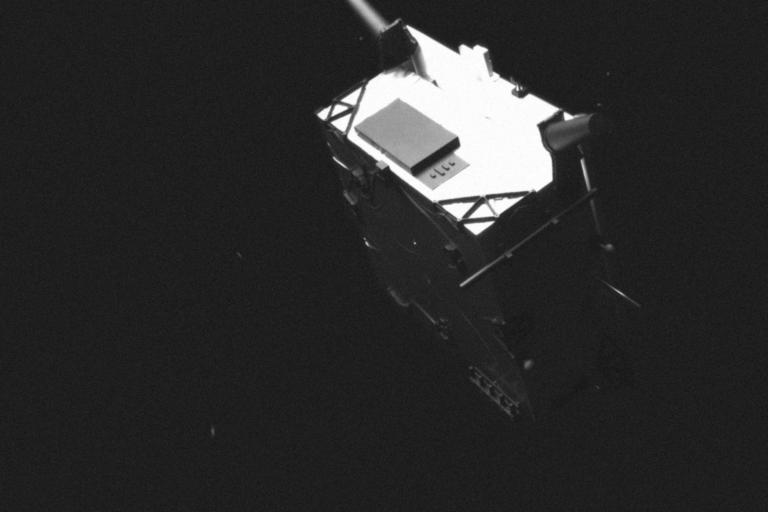}}; \\
      \node{\imgcell{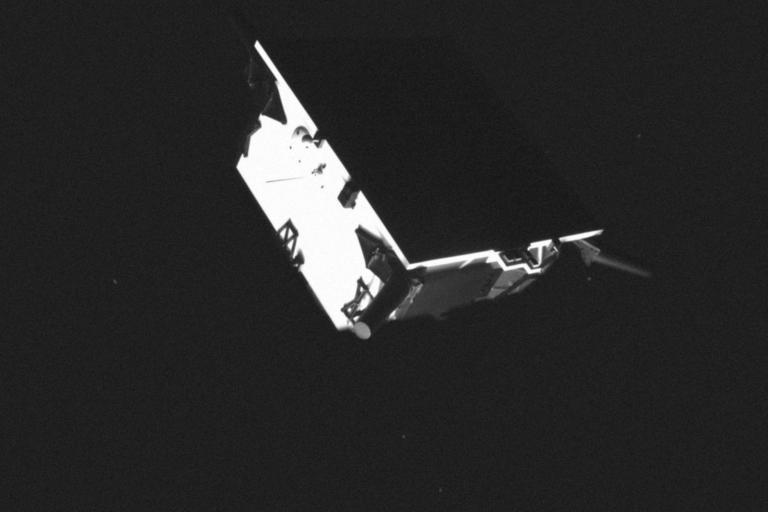}}; \\
      \node{\imgcell{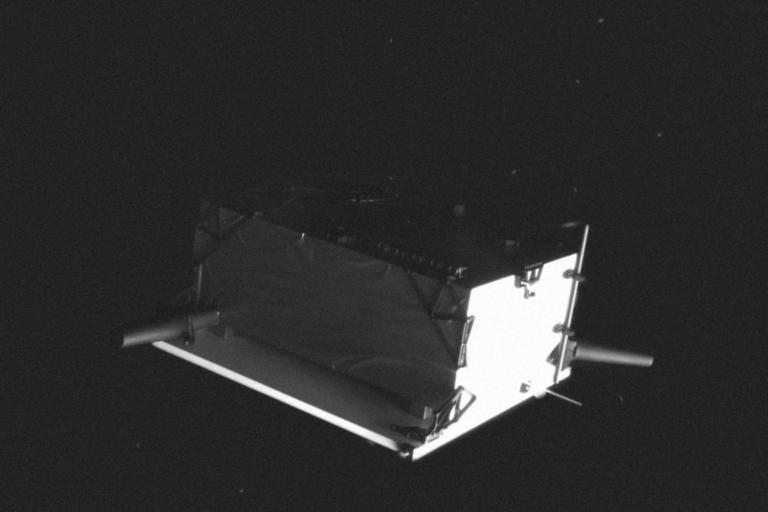}}; \\
    };
\end{pgfonlayer}
\node[block, fit=(grid1), inner sep=1pt, fill=black!25] (set1) {};
\node[color=black!80] (set1legend) at ($(set1.south west)+(-6pt,-8pt)$) {\normalsize $S_{\textnormal{\small Orig}}$};

\node[net, anchor=west, minimum width=18mm] (nerf)
  at ($(set1.east)+(18mm,0)$) {\normalsize NeRF $M_{\Phi}$};

\node (text1) [above left=4mm and 2mm of nerf.north, align=center] {Pose \\ labels};
\node (text2) [above right=4mm and -1mm of nerf.north, align=center] {App. Rand. \\ Config.};
\draw[arrow] (text1.south) to[bend left=10]  ($(nerf.north)+(-3mm,0)$);
\draw[arrow] (text2.south) to[bend right=10] ($(nerf.north)+(+3mm,0)$);

\draw[arrow] (set1.east) -- (nerf.west);

\coordinate (set2ULref) at ($(nerf.south west)+(0,-5mm)$); 

\begin{pgfonlayer}{infront}
\matrix (grid2) [matrix of nodes,
  nodes={inner sep=0pt, anchor=center},
  column sep=0pt, row sep=0pt,
  anchor=north west
] at ($(set2ULref)+(\setmargin,-\setmargin)$) {
  \node{\imgcell{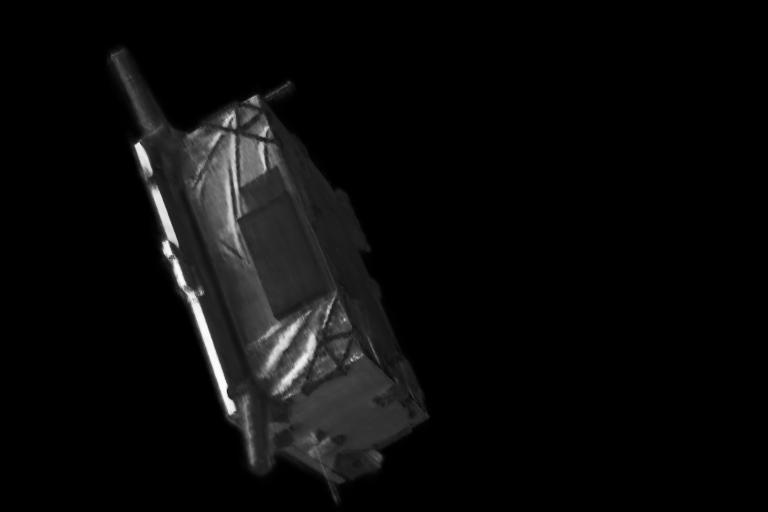}}; & \node{\imgcell{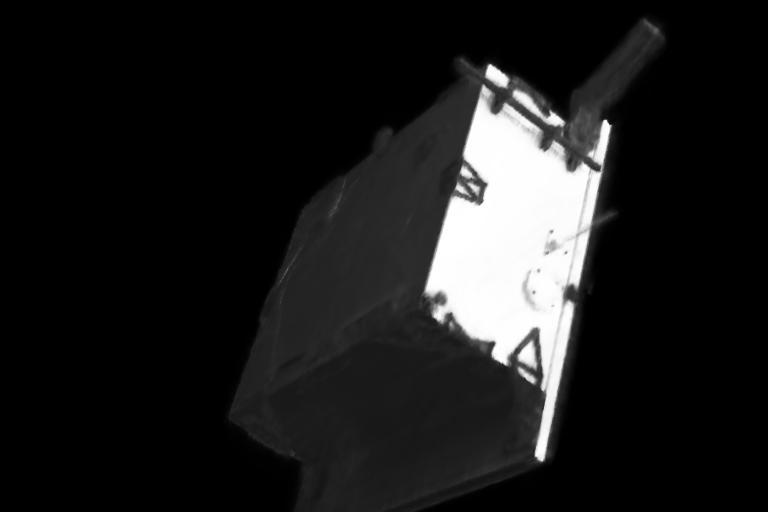}}; & \node{\imgcell{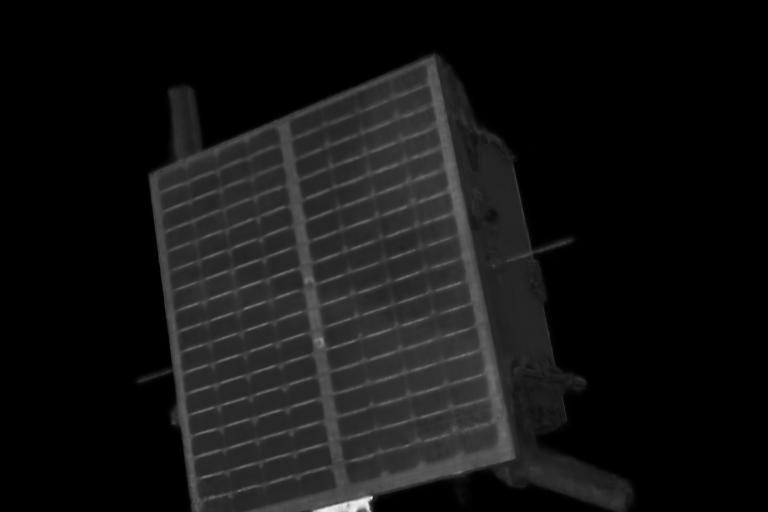}}; \\
  \node{\imgcell{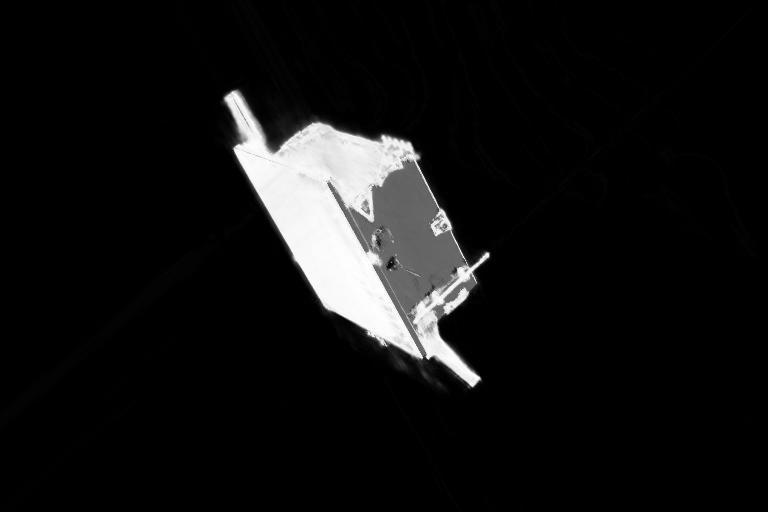}}; & \node{\imgcell{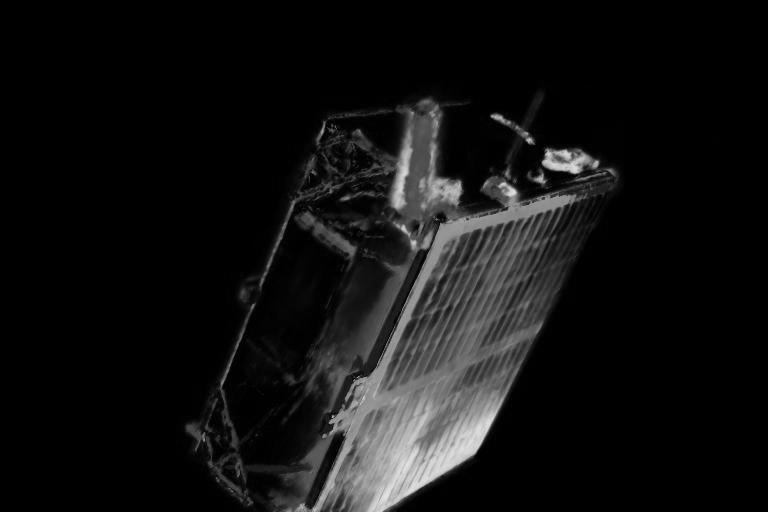}}; & \node{\imgcell{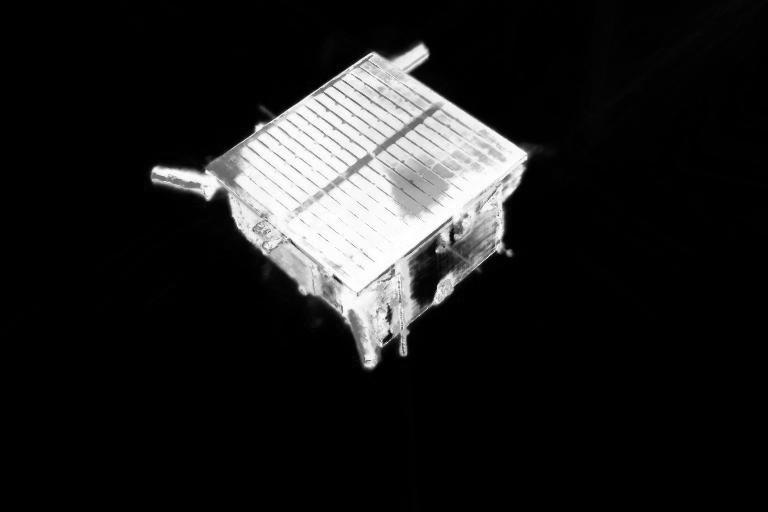}}; \\
  \node{\imgcell{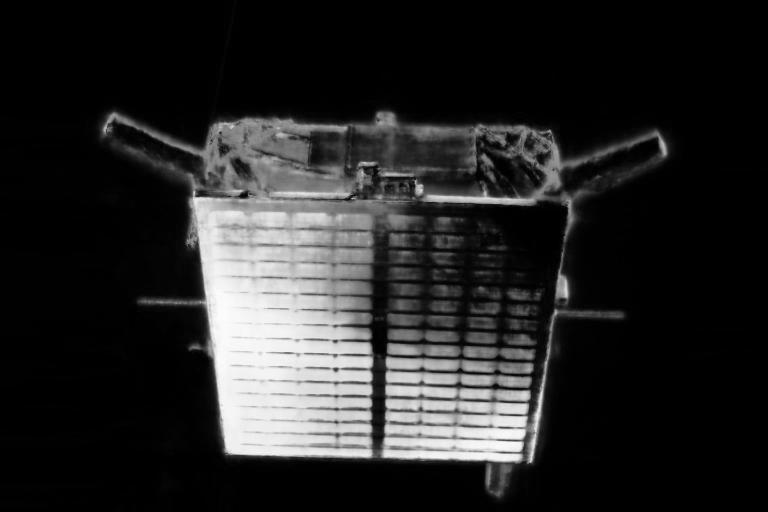}}; & \node{\imgcell{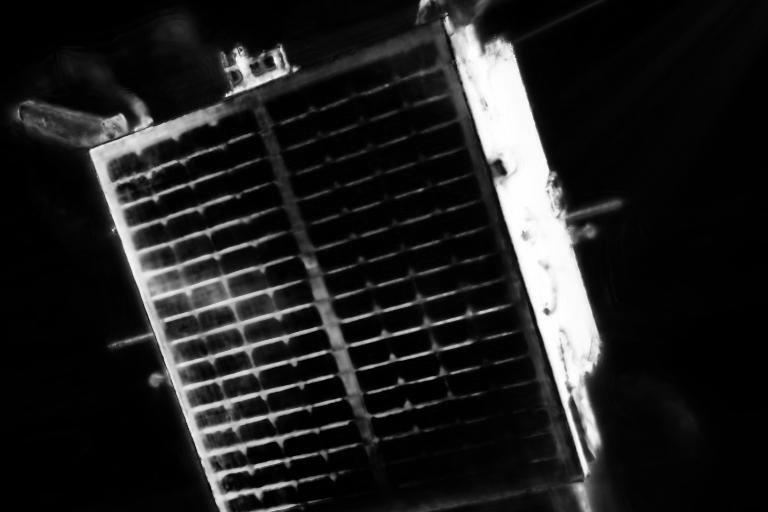}}; & \node{\imgcell{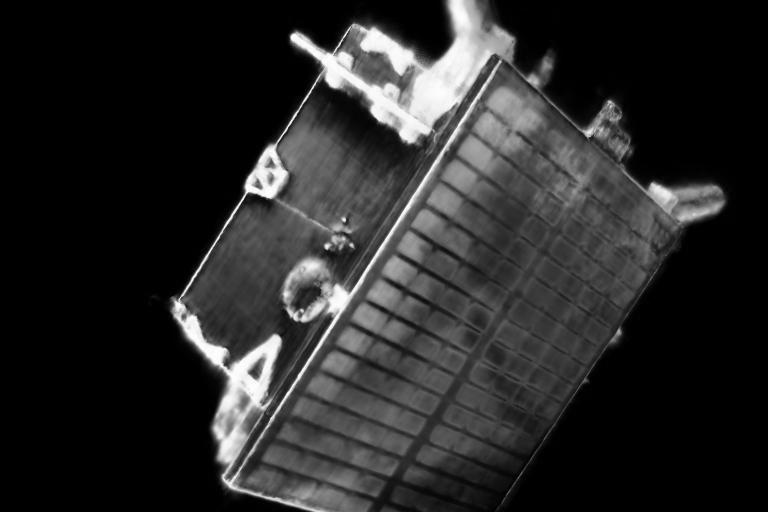}}; \\
};
\end{pgfonlayer}

\node[block, fit=(grid2), inner sep=1pt, fill=blue!25] (set2) {};
\node[color=blue!60] (set2legend) at ($(set2.north east)+(-18pt,8pt)$) {\large $S_{\textnormal{\small NeRF}}$};

\draw[arrow] (nerf.south) -- ($(set2.north -| nerf.south)$);

\node (union) [left=6mm of set2.west, anchor=east, font=\Large] {$\cup$};

\coordinate (set12union) at ($(set1.south east)+(-1mm,0mm)$); 
\draw[arrow] (set12union) |- (union);
\draw[arrow] (set2.west) -- (union);

\coordinate (set3ULref_vert) at ($(union.south)+(-0mm,-13mm)$); 
\coordinate (set3ULref) at ($(set3ULref_vert)+(-4mm,0mm)$); 

\begin{pgfonlayer}{infront}
    \matrix (grid3) [matrix of nodes,
      nodes={inner sep=0pt, anchor=center},
      column sep=0pt, row sep=0pt,
      anchor=north west
    ] at ($(set3ULref)+(2pt,-2pt)$) {
      \node{\imgcell{imgs/lightbox/img000201}}; & \node{\imgcell{imgs/lightbox/img000538}}; & \node{\imgcell{imgs/lightbox/img000437}}; \\
      \node{\imgcell{imgs/nerf/ill_rand/img000228}}; & \node{\imgcell{imgs/nerf/ill_rand/img000253}}; & \node{\imgcell{imgs/nerf/ill_rand/img000641}}; \\
      \node{\imgcell{imgs/nerf/color_mlp_05/img000644}}; & \node{\imgcell{imgs/nerf/color_mlp_05/img001033}}; & \node{\imgcell{imgs/nerf/color_mlp_05/img001516}}; \\
      \node{\imgcell{imgs/nerf/color_hid_10/img000011}}; & \node{\imgcell{imgs/nerf/color_hid_10/img000135}}; & \node{\imgcell{imgs/nerf/color_hid_10/img000412}}; \\
    };
\end{pgfonlayer}
\node[block, fit=(grid3), inner sep=1pt, fill=NavyBlue!50] (set3) {};
\node[color=NavyBlue!90] (set3legend) at ($(set3.north east)+(15pt,-8pt)$) {\large $S_{\textnormal{\small Augm}}$};

\draw[arrow] (union) -- (set3ULref_vert);

\newdimen\Xleft
\newdimen\Xright
\newdimen\Xcenter
\newdimen\Ypos


\coordinate (set1out) at ($(set1.south)+(4mm,0mm)$); 
\coordinate (set3out) at ($(set3.south west)+(4mm,0mm)$); 
\coordinate (posY) at ($(set3.south)+(0,-\posenetdrop)$);

\pgfextractx{\Xleft}{\pgfpointanchor{set1out}{south}}
\pgfextractx{\Xright}{\pgfpointanchor{set3out}{south}}
\pgfextracty{\Ypos}{\pgfpointanchor{posY}{center}}

\coordinate (tL) at (\Xleft,\Ypos);
\coordinate (tR) at (\Xright,\Ypos);

\draw[dblarrow gray] (set1out) -- (tL);     
\draw[dblarrow] (set3out) -- (tR);    

\Xcenter=\dimexpr(\Xleft+\Xright)/2\relax
\coordinate (posenet_center) at ($(\Xcenter,\Ypos)+(3mm,0mm)$); 
\node[net, anchor=north] (posenet)
  at (posenet_center) {\small Pose Estimation \\ \small Network $F_\Theta$};

\node (leftimg) [anchor=east] at ($(posenet.west)+(-4mm,0)$) {\includegraphics[width=11mm]{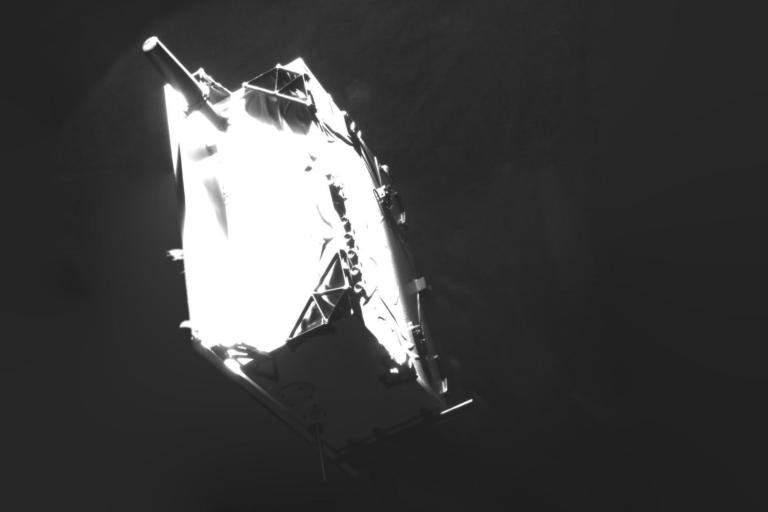}};
\node (leftlegend) [below=-2mm of leftimg] {\normalsize $I_{\textnormal{\footnotesize Test}}$};

\coordinate (arrow_basis) at ($(leftimg.east)+(-1mm,0mm)$);
\draw[arrow] (arrow_basis) -- (posenet.west);

\node (outtext) [right=8mm of posenet.east, anchor=west, align=center] {\normalsize 6D Pose \\ \large ($q$,$t$)};
\draw[arrow] (posenet.east) -- (outtext.west);

\begin{scope}[on background layer]
  \node[draw, rounded corners=4pt, thick,
        fill=ForestGreen!08,             
        inner sep=6.5pt,            
        fit=(set3)(set2)(set2legend)(nerf)(union)(text1)(text2)
  ] (bgblock) {};
  \node[anchor=north west, font=\small\bfseries, color=ForestGreen!60]
    at ($(bgblock.north)+(-68pt,12pt)$) {Our NeRF-based Image Augmentation};

    \node[draw, rounded corners=8pt, dotted, 
        fill=ForestGreen!25,             
        fit=(text1)(text2),
        thick,
        inner sep=1pt
      ] (rand_block) {};
  \node[anchor=south, font=\small, color=ForestGreen]
    at ($(rand_block.south east)+(7pt,-11pt)$) {Randomization};
\end{scope}

\end{tikzpicture}
\vspace{-0.3cm}
\caption{Our NeRF-based offline dataset augmentation method. Instead of training a pose estimation network on dataset $S_{\textnormal{\small Orig}}$ limited in terms of both volume and diversity, we train the network on an augmented set $S_{\textnormal{\small Augm}}$ that combines $S_{\textnormal{\small Orig}}$ with a set $S_{\textnormal{\small NeRF}}$. This set contains images generated through a NeRF trained on $S_{\textnormal{\small Orig}}$ under novel pose labels and randomized appearance conditions. This dataset augmentation enhances the pose estimator's generalization capabilities.}
\label{fig_taes_overview_sets}
\end{figure}

%% file: tikz_background.tex
    \begin{minipage}{.5\textwidth}
    \begin{flushleft}
    \begin{tikzpicture}[node distance=1.0cm, >=Stealth, align=left]
        \node (A) [rectangle, draw] {Pose ($\boldsymbol{q}$, $\boldsymbol{t}$)};
        \node (B) [rectangle, draw, below=of A] {{\small $W \!\times\! H$} rays};
        \node (C) [rectangle, draw, below=of B] {{\small$W \!\times\! H \!\times\! N$} samples};
        \node (D) [rectangle, draw, below=of C] {{\small$W \!\times \! H \! \times \! N$} points};
        \node (E) [rectangle, draw, below=of D] {Image $I$};
        
        
        \node (X3) [rectangle, draw] at ($(C)!0.5!(D) + (-1.9cm, 0)$) {\textcolor{green!45}{$\boldsymbol{e}$}};
        \draw[->] (X3.east) -- ($(C)!0.5!(D)$) node[midway, above, text width=3cm, align=center] {};

        \def\rectwidth{3.1}
        \def\rectheight{4.5}
        \begin{scope}[on background layer]
            \draw[rounded corners=5pt, fill=red!05, draw=none]
            ($(C) + (-\rectwidth/2, -\rectheight/2)$) 
            rectangle 
            ($(C) + (\rectwidth/2, \rectheight/2)$);
        \end{scope}
        
        \node (tempf) [left=0.8cm of B] {};
        \node (F) at ($(tempf)+(-0.1cm,0.45cm)$) [text=red!90, font=\Large\bfseries] {$M_\Phi$};
        
        \def\distbox{0.5cm}
        \def\distboxB{1.5cm}
        
        \draw[->] (A) -- (B) node[very near start, right, text width=8.75cm, anchor=north west] {
            \hspace{\distbox} $\mathcal{K}$: Projecting rays through {\small$W \!\times\! H$} image pixels\\\vspace{0.075cm}
            $ \hspace{\distboxB} \displaystyle (q,t) \xrightarrow{\mathcal{K}} \left\{ \left(c,d\right)_{ij}\right\}$
        };
        \draw[->] (B) -- (C) node[very near start, right, text width=8.75cm, anchor=north west]{
            \hspace{\distbox} $\mathcal{S}$: Sampling N points along each ray\\\vspace{0.075cm}
            $ \hspace{\distboxB} \displaystyle \left(c,d\right)_{ij} \xrightarrow{\mathcal{S}} \left\{ \left(x,y,z,\theta,\phi\right)_{ijk}\right\}$
        };
        \draw[->] (C) -- (D) node[very near start, right, text width=8.75cm, anchor=north west]{
            \hspace{\distbox} $\mathcal{F}$: Neural field inference\\\vspace{0.075cm}
            $ \hspace{\distboxB} \displaystyle \left(x,y,z,\theta,\phi\right)_{ijk}, \boldsymbol{e} \xrightarrow{\mathcal{F}} \left(r,g,b,\sigma\right)_{ijk}$
        };
        \draw[->] (D) -- (E) node[very near start, right, text width=8.75cm, anchor=north west] {
            \hspace{\distbox} $\mathcal{R}$: Differentiable ray tracing\\ \vspace{0.075cm}
            $ \hspace{\distboxB} \displaystyle \left\{ \left(r,g,b,\sigma\right)_{ijk}\right\} \xrightarrow{\mathcal{R}} I $
        };
    \end{tikzpicture}
    \end{flushleft}
    \end{minipage}
    \hfill    
    \begin{minipage}{.5\textwidth}
    \begin{flushright}
    \vspace{0.7cm}
    \begin{tikzpicture}[node distance=0.7cm and 1.0cm, auto, >=Stealth]
        \tikzstyle{block} = [draw, fill=blue!05, rounded corners, minimum height=3.0cm, minimum width=5.1cm]
        \tikzstyle{component} = [draw, fill=white, rounded corners, minimum height=0.9cm, minimum width=1.6cm]
        \tikzstyle{input} = [coordinate]
        \tikzstyle{output} = [coordinate]
        \tikzstyle{arrow} = [->, thick]
        
        \node[block] (main) {};   
        \node[input, above=of main, yshift=-0.4cm, xshift=0.35cm] (F) {};  
        
        \node[input, left=of main, xshift=0.6cm, yshift=0.9cm] (A) {};
        \node[input, left=of main, xshift=0.6cm, yshift=-0.2cm] (B) {};
        \node[input, left=of main, xshift=0.6cm, yshift=-0.9cm] (E) {}; 
        
        \node[output, right=of main, xshift=-0.6cm, yshift=0.9cm] (C) {};
        \node[output, right=of main, xshift=-0.6cm, yshift=-0.9cm] (D) {};
        
        \node[component, right=0.75cm of A] (comp1) {Pos. Enc.};
        \node[component, below=0.2cm of comp1] (comp2) {Dir. Enc.};
        \node[component, right=0.75cm of comp1] (comp3) {Density Field};
        \node[component, below=0.9cm of comp3] (comp4) {Color Field};
        
        \node[left=0cm of F] {{\textcolor{blue!45}{\Large $\mathcal{F}$}}};
        \node[left=0cm of A] {($x$,$y$,$z$)};
        \node[left=0cm of B] {($\theta$,$\phi$)};
        \node[left=0cm of E] {\textcolor{green!45}{$\boldsymbol{e}$}};
        \node[right=0cm of C] {$\sigma$};
        \node[right=0cm of D] {($r$,$g$,$b$)};
        
        \node (tempa) [right=0.3cm of comp1] {};
        \node (a) at ($(tempa)+(0cm,-0.35cm)$) {$\boldsymbol{F_{\textnormal{pos}}}$};
        \node (tempb) [right=0.3cm of comp2] {};
        \node (b) at ($(tempb)+(0cm,0.0cm)$) {$\boldsymbol{F_{\textnormal{dir}}}$};
        \node (tempc) [below=0.3cm of comp3] {};
        \node (c) at ($(tempc)+(0.25cm,0cm)$) {$\boldsymbol{F_{\sigma}}$};
        
        \draw[arrow] (A) -- (comp1);
        \draw[arrow] (B) -- (comp2);
        \draw[arrow] (comp1) -- (comp3);
        \draw[arrow] (comp3) -- (C);
        \draw[arrow] (comp2) -- (comp4);
        \draw[arrow] (comp3) -- ++(0,-0.5) -| (comp4);
        \draw[arrow] (E) -- (comp4); 
        \draw[arrow] (comp4) -- (D);
    \end{tikzpicture}
    \end{flushright}
    \end{minipage}

%% file: tikz_figure_method.tex
\begin{figure*}[t]
    \centering

    \newcommand{\SubFigWidth}{0.97\linewidth}  
    \newcommand{\SubFigVGap}{0.2cm}           
    \newcommand{\OuterPadding}{0.3cm}          
    
    \newcommand{\CornerRadius}{4pt}     
    \newcommand{\InnerLineWidth}{0.8pt} 
    \newcommand{\OuterLineWidth}{1.1pt}        

    \newcommand{\ArrowLineWidth}{0.8pt} 
    \newcommand{\ArrowTip}{Stealth}     
    
    \newlength{\ShortArrowLen}
    \setlength{\ShortArrowLen}{6mm}
    \newlength{\LongArrowLen}
    \setlength{\LongArrowLen}{15mm}
    
    \newlength{\StageOneMinW}  \setlength{\StageOneMinW}{20mm}
    \newlength{\StageOneMinH}  \setlength{\StageOneMinH}{12mm}
    
    \newlength{\StageTwoMinW}  \setlength{\StageTwoMinW}{18mm}
    \newlength{\StageTwoMinH}  \setlength{\StageTwoMinH}{14mm}
    
    \newlength{\StageThreeMinW}\setlength{\StageThreeMinW}{20mm}
    \newlength{\StageThreeMinH}\setlength{\StageThreeMinH}{17mm}
    
    \newlength{\ComponentMinW} \setlength{\ComponentMinW}{10mm}
    \newlength{\ComponentMinH} \setlength{\ComponentMinH}{10mm}
    
    \newlength{\ComponentsthreeMinW} \setlength{\ComponentsthreeMinW}{16mm}
    \newlength{\ComponentsthreeMinH} \setlength{\ComponentsthreeMinH}{11mm}
    
    \begin{tikzpicture}[every node/.style={inner sep=6pt}]
        \tikzset{
            subfig/.style={
                draw,
                rounded corners=\CornerRadius,
                line width=\InnerLineWidth,
                minimum width=\SubFigWidth,
                align=left,
                anchor=north
            },
            toplabel/.style={
                anchor=south west,
                inner sep=4pt,
                font=\bfseries
            },
            blockbase/.style={
                draw,
                rounded corners=\CornerRadius,
                line width=\InnerLineWidth,
                align=center,
                inner sep=4pt
            },
            stage1block/.style={
                blockbase,
                minimum width=\StageOneMinW,
                minimum height=\StageOneMinH
            },
            stage2block/.style={
                blockbase,
                minimum width=\StageTwoMinW,
                minimum height=\StageTwoMinH
            },
            stage3block/.style={
                blockbase,
                minimum width=\StageThreeMinW,
                minimum height=\StageThreeMinH
            },
            componentblock/.style={
                blockbase,
                minimum width=\ComponentMinW,
                minimum height=\ComponentMinH
            },
            componentblocksthree/.style={
                blockbase,
                minimum width=\ComponentsthreeMinW,
                minimum height=\ComponentsthreeMinH,
                inner sep = 0pt
            },
            shortarrow/.style={
                -{Stealth[length=3mm, width=2mm]},
                line width=0.8pt
            },
            longarrow/.style={
                -{Stealth[length=4mm, width=2.5mm]},
                line width=0.8pt
            },
            doublearrow/.style={
                double,                         
                line width=0.8pt,               
                double distance=3pt,            
                {Bar[width=0.8pt]}-{Stealth[length=10pt,width=10pt]}, 
                line cap=butt                   
            },
            lossbox/.style={
                draw,
                rounded corners=2pt,
                minimum width=12mm,
                minimum height=8mm,
                fill=blue!05,
                inner sep=2pt
            }
        }

    \def\StageOneH{1.7cm}
    \def\StageTwoH{6.2cm}
    \def\StageThreeH{4.8cm}

    \begin{scope}[on background layer]
        
        \node[subfig, fill=green!1.0!white, minimum height=\StageTwoH] (stage2) {};
        
        \node[subfig, fill=green!1.0!white, minimum height=\StageThreeH, below=\SubFigVGap of stage2] (stage3) {};
    \end{scope}
    
    \def\sTwoXSep{8mm}          
    \def\branchDY{12mm}         
    \def\exprDX{5mm}            
    \def\exprDY{4mm}            
    \def\toCompDX{29mm}         
    \def\toCompDXR{20mm}         
    \def\lossYShift{4mm}        
    \def\labelExtraSepX{8mm}    
    
    \node[stage2block, anchor=west, font=\small] (s2_block1) at ([xshift=2mm, yshift=2.0mm]stage2.west) {Pose \& \\ Illumination \\ Correction};
    
    \node[anchor=south] (s2_math1) at ([yshift=4mm]s2_block1.north) {$\left(\hat{c}, \hat{d}\right)$};
    
    \node[anchor=west] (s2_math2) at ([xshift=\sTwoXSep]s2_block1.east) {$\left(\tilde{c}, \tilde{d}, e \right)$};

    \draw[shortarrow] (s2_math1.south) -- (s2_block1.north);    
    \draw[longarrow] (s2_block1.east) -- (s2_math2.west);

    \node[componentblock, anchor=west] (s2_c1_top) at ([xshift=\sTwoXSep, yshift=\branchDY] s2_math2.east) {$\mathcal{S}_{\Phi}$};
    
    \node[anchor=west, font=\small] (s2_m1_top_up) at ([xshift=\exprDX, yshift=\exprDY] s2_c1_top.east) {$\left\{\left(x,y,z\right)_n\right\}$};
    \node[anchor=west, font=\small] (s2_m1_top_dn) at ([xshift=\exprDX, yshift=-\exprDY] s2_c1_top.east) {$\left\{\left(\theta,\phi\right)_n\right\}$};
    
    \node[componentblock, anchor=west] (s2_b2_top) at ([xshift=\toCompDX] s2_c1_top.east) {$\mathcal{F}_{\Phi}$};
    
    \node[anchor=west, font=\small] (s2_m2_top_up) at ([xshift=\exprDX, yshift=\exprDY] s2_b2_top.east) {$\left\{y_n\right\}$};
    \node[anchor=west, font=\small] (s2_m2_top_dn) at ([xshift=\exprDX, yshift=-\exprDY] s2_b2_top.east) {$\left\{\sigma_n\right\}$};
    
    \node[componentblock, anchor=west] (s2_c3_top) at ([xshift=\toCompDXR] s2_b2_top.east) {$\mathcal{R}$};
        
    \node[anchor=west] (s2_m3_top) at ([xshift=\sTwoXSep] s2_c3_top.east) {$\hat{I}$};

    \draw[longarrow] (s2_math2.east) -- (s2_c1_top.west);
    \draw[shortarrow] (s2_c1_top.east) -- (s2_m1_top_up.west);
    \draw[shortarrow] (s2_c1_top.east) -- (s2_m1_top_dn.west);
    \draw[shortarrow] (s2_m1_top_up.east) -- (s2_b2_top.west);
    \draw[shortarrow] (s2_m1_top_dn.east) -- (s2_b2_top.west);
    \draw[shortarrow] (s2_b2_top.east) -- (s2_m2_top_up.west);
    \draw[shortarrow] (s2_b2_top.east) -- (s2_m2_top_dn.west);
    \draw[shortarrow] (s2_m2_top_up.east) -- (s2_c3_top.west);
    \draw[shortarrow] (s2_m2_top_dn.east) -- (s2_c3_top.west);
    \draw[shortarrow] (s2_c3_top.east) -- (s2_m3_top.west);

    \begin{scope}[on background layer]
        \node[draw, rounded corners, inner sep=6pt, fit=(s2_c1_top) (s2_c3_top), fill=red!10!white] (branchtop)  {};
    \end{scope}

    \node[above left=-8mm and -1mm of branchtop.north west, text=red!75!black, font=\large\bfseries] (branchtoplegend) {$M_{\Phi}$};

    \node[lossbox, anchor=south, text=red!75!black, fill=red!10!white] (s2_box_top)at ([yshift=\lossYShift] s2_m3_top.north) {$\mathcal{L_{\textnormal{Photo}}}$};
    
    \draw[shortarrow] (s2_m3_top.north) -- (s2_box_top.south);
    \draw[doublearrow] (s2_box_top.west) -- ($(s2_box_top.west -| s2_block1.west)$);
    
    \node[anchor=west] (s2_l_top) at ([xshift=\labelExtraSepX] s2_box_top.east) {$I$};
    \draw[shortarrow] (s2_l_top.west) -- (s2_box_top.east);

    \node[componentblock, anchor=west] (s2_c1_bot) at ([xshift=\sTwoXSep, yshift=-\branchDY] s2_math2.east) {$\mathcal{S}_{\Psi}$};
    
    \node[anchor=west, font=\small] (s2_m1_bot_up) at ([xshift=\exprDX, yshift=\exprDY] s2_c1_bot.east) {$\left\{\left(x,y,z\right)_n\right\}$};
    \node[anchor=west, font=\small] (s2_m1_bot_dn) at ([xshift=\exprDX, yshift=-\exprDY] s2_c1_bot.east) {$\left\{\left(\theta,\phi\right)_n\right\}$};
    
    \node[componentblock, anchor=west] (s2_b2_bot) at ([xshift=\toCompDX] s2_c1_bot.east) {$\mathcal{F}_{\Psi}$};
    
    \node[anchor=west, font=\small] (s2_m2_bot_up) at ([xshift=\exprDX, yshift=\exprDY] s2_b2_bot.east) {$\left\{y_n\right\}$};
    \node[anchor=west, font=\small] (s2_m2_bot_dn) at ([xshift=\exprDX, yshift=-\exprDY] s2_b2_bot.east) {$\left\{\sigma_n\right\}$};
    
    \node[componentblock, anchor=west] (s2_c3_bot) at ([xshift=\toCompDXR] s2_b2_bot.east) {$\mathcal{R}$};
    
    \node[anchor=west] (s2_m3_bot) at ([xshift=\sTwoXSep] s2_c3_bot.east) {$\hat{I}$};

    \draw[longarrow] (s2_math2.east) -- (s2_c1_bot.west);
    \draw[shortarrow] (s2_c1_bot.east) -- (s2_m1_bot_up.west);
    \draw[shortarrow] (s2_c1_bot.east) -- (s2_m1_bot_dn.west);
    \draw[shortarrow] (s2_m1_bot_up.east) -- (s2_b2_bot.west);
    \draw[shortarrow] (s2_m1_bot_dn.east) -- (s2_b2_bot.west);
    \draw[shortarrow] (s2_b2_bot.east) -- (s2_m2_bot_up.west);
    \draw[shortarrow] (s2_b2_bot.east) -- (s2_m2_bot_dn.west);
    \draw[shortarrow] (s2_m2_bot_up.east) -- (s2_c3_bot.west);
    \draw[shortarrow] (s2_m2_bot_dn.east) -- (s2_c3_bot.west);
    \draw[shortarrow] (s2_c3_bot.east) -- (s2_m3_bot.west);

    \begin{scope}[on background layer]
        \node[draw, rounded corners, inner sep=6pt, fit=(s2_c1_bot) (s2_c3_bot), fill=blue!10!white] (branchbot) {};
    \end{scope}
    \node[above left=2mm and -1mm of branchbot.south west, text=blue!75!black, font=\large\bfseries] (branchbotlegend) {$M_{\Psi}$};

    \node[lossbox, anchor=south, text=blue!75!black, fill=blue!10!white] (s2_box_bot1) at ([yshift=\lossYShift] s2_m3_bot.north) {$\mathcal{L_{\textnormal{Photo}}}$};
    \draw[shortarrow] (s2_m3_bot.north) -- (s2_box_bot1.south);    
    \draw[doublearrow] (s2_box_bot1.west) -- ($(s2_box_bot1.west -| s2_c1_bot.west)$);

    \node[lossbox, anchor=north, text=blue!75!black, fill=blue!10!white] (s2_box_bot2) at ([yshift=-\lossYShift] s2_m2_bot_dn.south) {$\mathcal{L_{\sigma}}$};
    \draw[shortarrow] (s2_m2_bot_dn.south) -- (s2_box_bot2.north);    
    \draw[doublearrow] (s2_box_bot2.west) -- ($(s2_box_bot2.west -| s2_c1_bot.west)$);

    \draw[shortarrow] (s2_l_top.west) -- (s2_box_bot1.east);
    
    \node[anchor=west] (s2_l_bot2) at ([xshift=\labelExtraSepX] s2_box_bot1.east) {$\hat{m}$};
    \draw[shortarrow] (s2_l_bot2.south) |- (s2_box_bot2.east);   

    \draw[shortarrow] (s2_l_bot2.west) -- (s2_box_top.east);
    \draw[shortarrow] (s2_l_bot2.west) -- (s2_box_bot1.east);

    \def\sthreeDX{22mm}          
    \def\sthreeBlockGap{9mm}     
    \def\sthreeExprDX{8mm}      

    \node (stage3ref) at ([xshift=\sthreeDX, yshift=2mm] stage3.west)  {};
    
    \node[stage3block, anchor=west, text=blue!75!black, fill=blue!10!white] (s3_btop) at ([yshift=\sthreeBlockGap] stage3ref) {$\mathcal{M}_{\Psi}$};
    \node[stage3block, anchor=west, fill=red!10!white] (s3_bbot) at ([yshift=-\sthreeBlockGap] stage3ref) {Sampling \\ $f^{\sigma}_{\phi}$ Inf.};
    
    \coordinate (s3_mid_west) at ($ (s3_btop.west)!0.5!(s3_bbot.west) $);
    \node[anchor=west] (s3_math1) at (stage3.west |- s3_mid_west) {$(q,t)$};
    
    \def\sthreeLeadHX{2mm} 
    \draw[longarrow] (s3_math1.east) -- ++(\sthreeLeadHX,0) |- (s3_btop.west);
    \draw[longarrow] (s3_math1.east) -- ++(\sthreeLeadHX,0) |- (s3_bbot.west);
    
    \node[anchor=west] (s3_mtop) at ([xshift=\sthreeExprDX] s3_btop.east) {$m$};
    \draw[shortarrow] (s3_btop.east) -- (s3_mtop.west);
    
    \coordinate (s3_bbot_e_1) at ($ (s3_bbot.north east)!1/5!(s3_bbot.south east) $);
    \coordinate (s3_bbot_e_2) at ($ (s3_bbot.north east)!2/5!(s3_bbot.south east) $);
    \coordinate (s3_bbot_e_3) at ($ (s3_bbot.north east)!4/5!(s3_bbot.south east) $);
    
    \node[anchor=west] (s3_mbot1) at ([xshift=\sthreeExprDX] s3_bbot_e_1) {$F_{\textnormal{Dir}}$};
    \node[anchor=west] (s3_mbot2) at ([xshift=\sthreeExprDX] s3_bbot_e_2) {$F_{\sigma}$};
    \node[anchor=west] (s3_mbot3) at ([xshift=\sthreeExprDX] s3_bbot_e_3) {$\sigma$};
    
    \draw[shortarrow] (s3_bbot_e_1) -- (s3_mbot1.west);
    \draw[shortarrow] (s3_bbot_e_2) -- (s3_mbot2.west);
    \draw[shortarrow] (s3_bbot_e_3) -- (s3_mbot3.west);

    
    \def\sCompDX{22mm}        
    \def\sDashBoxY{5mm}       
    \def\sDashExprDX{8mm}     
    \def\sLeadHX{4mm}         
    \def\sOutDX{6mm}          
    
    \coordinate (s3_comp_w_mid) at ($(s3_mbot1)+(\sCompDX,0)$);
    
    \coordinate (s3_in2_proj) at ($(s3_comp_w_mid |- s3_mbot2)$);
    
    \coordinate (s3_west_north) at ($(s3_comp_w_mid)! 2 ! (s3_in2_proj)$);
    \coordinate (s3_west_south) at ($(s3_comp_w_mid)!-2 ! (s3_in2_proj)$);
    
    \coordinate (s3_east_north) at ($(s3_west_north)+(\ComponentsthreeMinW,0)$);
    \coordinate (s3_east_south) at ($(s3_west_south)+(\ComponentsthreeMinW,0)$);
    
    \node[componentblocksthree, inner sep=0mm, fit=(s3_west_south)(s3_west_north)(s3_east_south)(s3_east_north), fill=red!10!white] (s3_comp) {$f^{\textnormal{(col)}}_{\phi^{\textnormal{(col)}}+\epsilon^{(k)}}$};
    
    \coordinate (s3_comp_w_1) at ($(s3_west_south)!1/4!(s3_west_north)$);
    \coordinate (s3_comp_w_2) at ($(s3_west_south)!2/4!(s3_west_north)$);
    \coordinate (s3_comp_w_3) at ($(s3_west_south)!3/4!(s3_west_north)$);
    
    \draw[shortarrow] (s3_mbot1.east) -- (s3_comp_w_2);
    \draw[shortarrow] (s3_mbot2.east) -- (s3_comp_w_3);
    
    
    \node[anchor=south east,text=green!50!black] (s3_box_expr1) at ([yshift=\sDashBoxY, xshift=-\sDashExprDX] s3_comp.north) {$e^{(k)}$};
    
    \node[anchor=south,text=green!50!black] (s3_box_expr2) at ([yshift=\sDashBoxY] s3_comp.north) {$\epsilon^{(k)}$};
    
    \node[
      draw, dashed, rounded corners=2pt, inner sep=2pt,
      fit=(s3_box_expr1)(s3_box_expr2),
      color=green!50!black
    ] (s3_dbox) {};
    \node[anchor=west,text=green!50!black] (s3_ctrl_text) at ([xshift=-1mm] s3_dbox.east) {\small{$\sim$ Random}};
    
    \draw[shortarrow] (s3_box_expr1.south) |- (s3_comp_w_1);
    
    \draw[shortarrow] (s3_box_expr2.south) -- (s3_comp.north);
    
    
    \node[anchor=west] (s3_comp_out)at ([xshift=\sOutDX] s3_comp.east) {$y^{(k)}$};
    
    \draw[shortarrow] (s3_comp.east) -- (s3_comp_out.west);
    

    \def\sLastCompDX{54mm}   
    
    \def\sLastCompH{12mm}
    
    \coordinate (s3_last_w_mid) at ($(s3_mbot3.east)+(\sLastCompDX,0)$);
    \coordinate (s3_last_aux_top) at ($(s3_last_w_mid)+(0,\sLastCompH)$);
    
    \coordinate (s3_last_w_top)    at ($ (s3_last_w_mid)!2/3!(s3_last_aux_top) $);
    \coordinate (s3_last_w_bottom) at ($ (s3_last_w_mid)!-1/3!(s3_last_aux_top) $);
    
    \coordinate (s3_last_e_top)    at ($(s3_last_w_top)+(\ComponentMinW,0)$);
    \coordinate (s3_last_e_bottom) at ($(s3_last_w_bottom)+(\ComponentMinW,0)$);
    
    \node[componentblock, inner sep=0mm, fit=(s3_last_w_bottom)(s3_last_w_top)(s3_last_e_bottom)(s3_last_e_top), fill=red!10!white] (s3_last_comp) {$\mathcal{R}$};
    
    \coordinate (s3_last_w_1_3) at ($ (s3_last_w_top)!1/3!(s3_last_w_bottom) $);
    \coordinate (s3_last_w_2_3) at ($ (s3_last_w_top)!2/3!(s3_last_w_bottom) $);
    
    \draw[shortarrow] (s3_mbot3.east) -- (s3_last_w_2_3);
    \draw[shortarrow] (s3_comp_out.south) |- (s3_last_w_1_3);
    
    \def\sLastOutDX{8mm}
    \node[anchor=west] (s3_last_expr1) at ([xshift=\sLastOutDX] s3_last_comp.east) {$I^{(k)}$};
    \draw[shortarrow] (s3_last_comp.east) -- (s3_last_expr1.west);
    
    \node[anchor=west] (s3_last_expr2) at ([xshift=\sLastOutDX] s3_last_expr1.east) {$\{I^{(k)}\}_{k=0}^{N_{\textnormal{cfg}}}$};
    \draw[shortarrow] (s3_last_expr1.east) -- (s3_last_expr2.west);

    \def\sContainerPad{6pt}
    
    \begin{scope}[on background layer]
        \node[
          draw,
          fill=green!5.0!white,
          rounded corners=\CornerRadius,
          line width=\InnerLineWidth,
          fit=(s3_last_expr1)(s3_last_comp)(s3_dbox),
          inner sep=\sContainerPad
        ] (mycontainer) {};
    \end{scope}
    
    \node[anchor=north east, inner sep=3pt,text=green!50!black] at (mycontainer.north east) {$0<k<N_{\textnormal{cfg}}$};

    \node[toplabel] at (stage2.south west) {1° Radiance Field Learning};
    \node[toplabel] at (stage3.south west) {2° Appearance-randomized Foreground Image Synthesis};

        \node[
            draw,
            rounded corners=\CornerRadius,
            line width=\OuterLineWidth,
            fit=(stage2)(stage3),
            inner sep=\OuterPadding
        ] (outerframe) {};

    \end{tikzpicture}

    \caption{Method Overview. We first learn two neural radiance fields $M_\Phi$ and $M_\Psi$ on the available images. $M_\Phi$ (in red) is supervised using a photometric loss and aims at accurately representing the target appearance. $M_\Psi$ (in blue) is supervised on the photometric loss as well as on its density field to enforce the learning of an accurate geometry representation. Then, we generate novel training samples using both representations. For each pose label ($q$, $t$), we generate a segmentation mask $m$ using $M_\Psi$ and $N_{\textnormal{cfg}}$ foreground images $I^{\textnormal{(k)}}$ corresponding to that pose through $M_\Phi$. To generate images exhibiting diverse appearance, we randomize either the illumination or the color (in green). The illumination is randomized through the appearance embedding $\mathbf{e}^{\textnormal{(k)}}$ while the color is augmented through the randomization of the color MLP's weights $\epsilon^{\textnormal{(k)}}$.}
    \label{fig_taes_method}
\end{figure*}

%% file: tikz_spnv2.tex
\pgfdeclarelayer{foreground}
\pgfsetlayers{background,main,foreground}
\begin{tikzpicture}[
    block/.style={draw, opacity=1.0, fill=white, shape=trapezium,
    trapezium stretches=true,
    trapezium left angle=110, trapezium right angle=-70, 
    shape border rotate=90, thick, rounded corners, minimum height=1.1cm, minimum width=2.0cm, align=center},
    head/.style={draw, thick, rounded corners, minimum height=1.0cm, minimum width=2.4cm, align=center},
    feature/.style={minimum height=0.5cm, minimum width=0.8cm, align=center},
    featurerawbox/.style={fill=red!2, draw=red!50, rounded corners},
    featurebox/.style={fill=green!2, draw=green!50, rounded corners},
    headsbox/.style={fill=blue!2, draw=blue!60, rounded corners},
    fuseTD/.style={draw=black!05, line width=0.4pt, -{Latex[length=3.5pt]}}, 
      fuseBU/.style={draw=black!05, line width=0.4pt, -{Latex[length=3.5pt]}}, 
      axisarrow/.style={draw=gray!20, line width=2.9pt, -{Latex[length=9pt]}}
]

\node (input)[inner sep=0, outer sep=0] {\includegraphics[width=2.0cm]{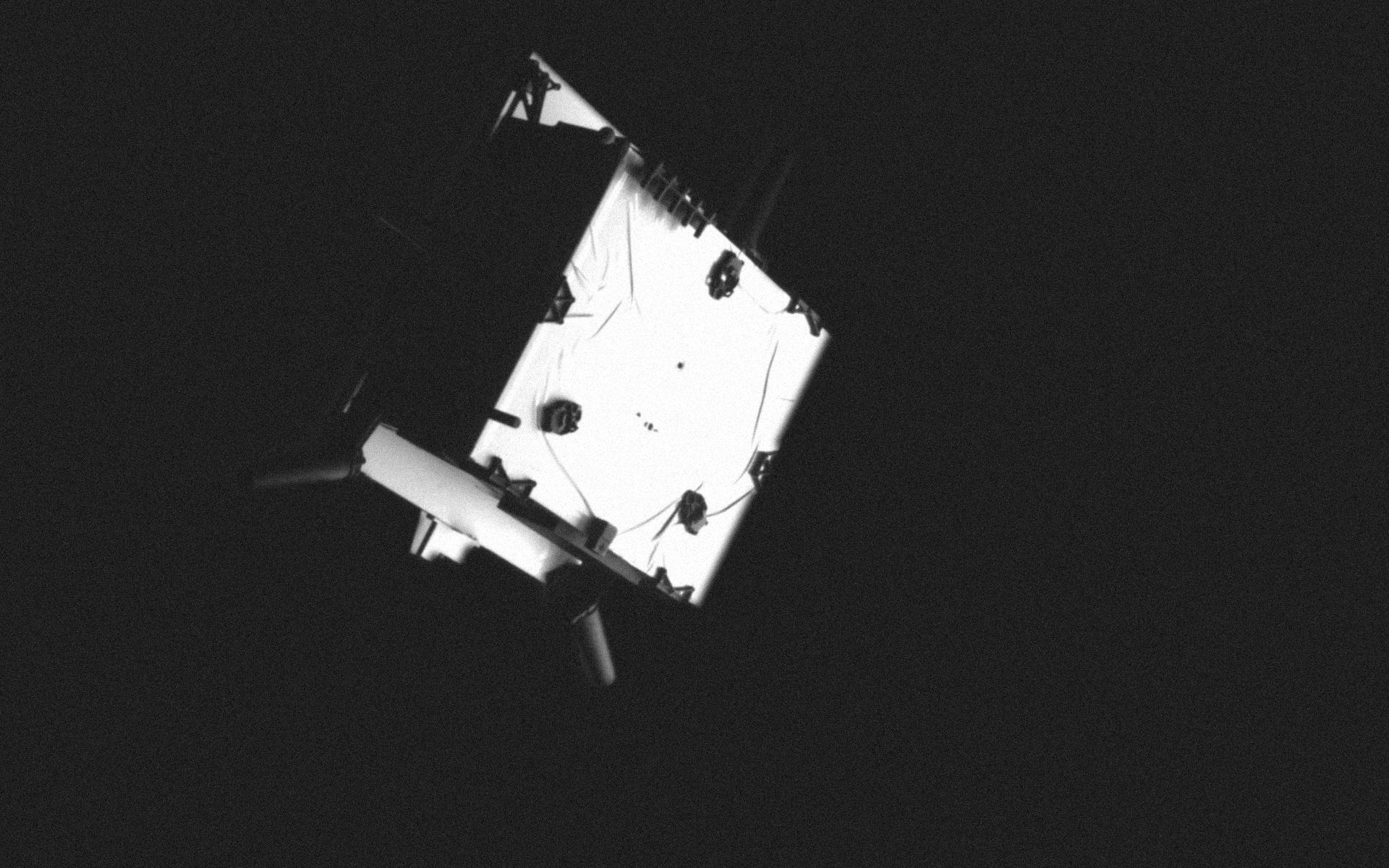}};
\node[above=-0.5pt of input, font=\small] {\small$I_{\scriptscriptstyle W \times H}$};

\begin{pgfonlayer}{foreground}
    \node[draw, minimum width=1.2cm, minimum height=2.2cm, align=center, rounded corners, below=0.3cm of input] (effnet) { \small EfficientNet \\ \small + BiFPN \\ \small Backbone};
\end{pgfonlayer}

\draw[->] (input) -- (effnet);

\coordinate (N) at ($(effnet.east)+(2.0cm,-1.1cm)$);
\node[head, above=0.0cm of N] (segm) {\small Segmentation};
\node[head, above=1.4cm of N] (heatmap) {\small Heatmap};
\node[head, above=2.8cm of N] (pose) {\small EfficientPose};

\coordinate (tapcolR) at ($(pose.west)+(-0.4cm,0)$);

\draw[->] (effnet.east) -- (tapcolR |- effnet.east) |- (heatmap.west);
\draw[->] (effnet.east) -- (tapcolR |- effnet.east) |- (segm.west);
\draw[->] (effnet.east) -- (tapcolR |- effnet.east) |- (pose.west);

\node[right=0.6cm of heatmap.east] (output_heat) {\includegraphics[height=1.2cm]{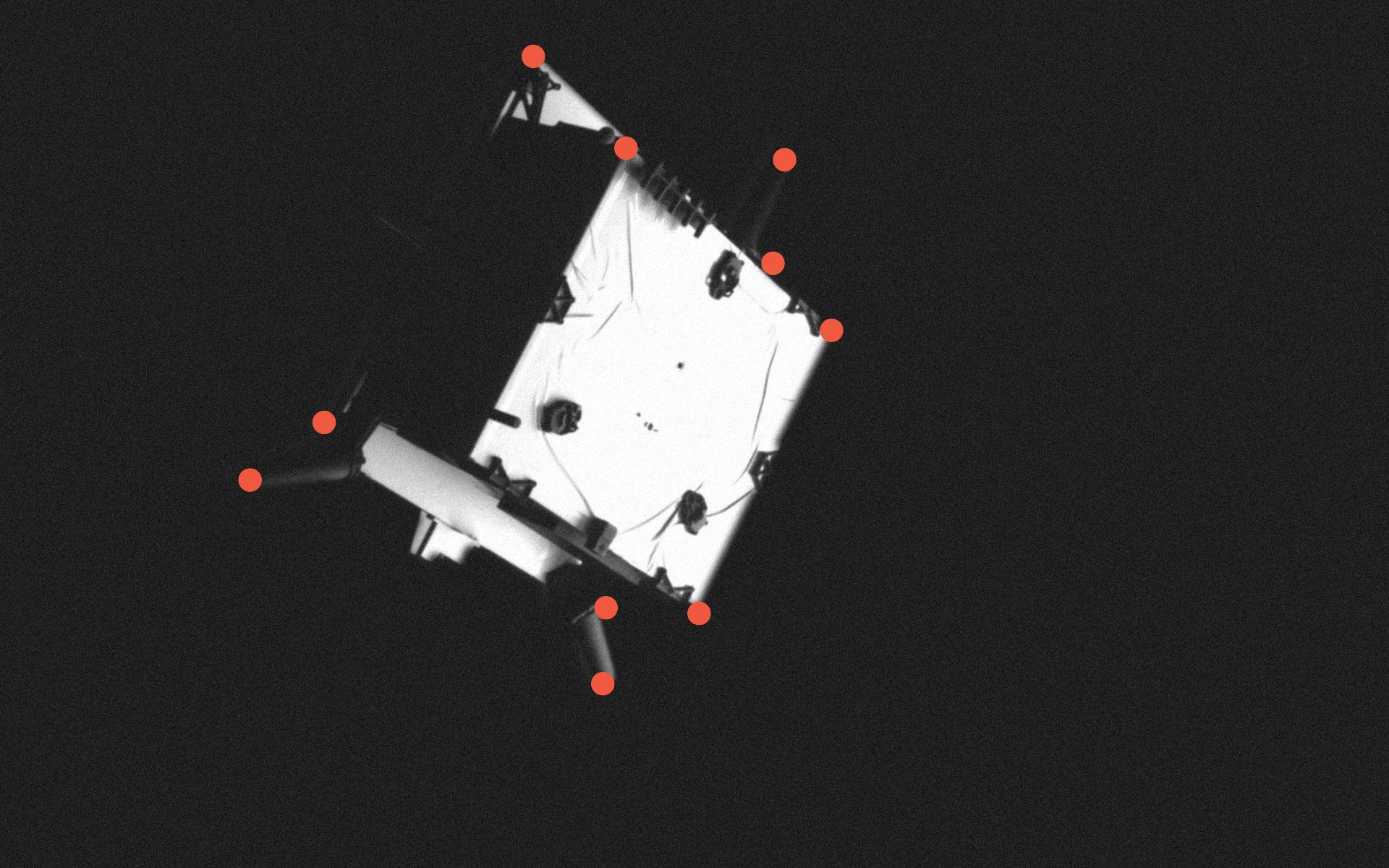}};
\node[right=0.6cm of segm.east] (output_segm) {\includegraphics[height=1.2cm]{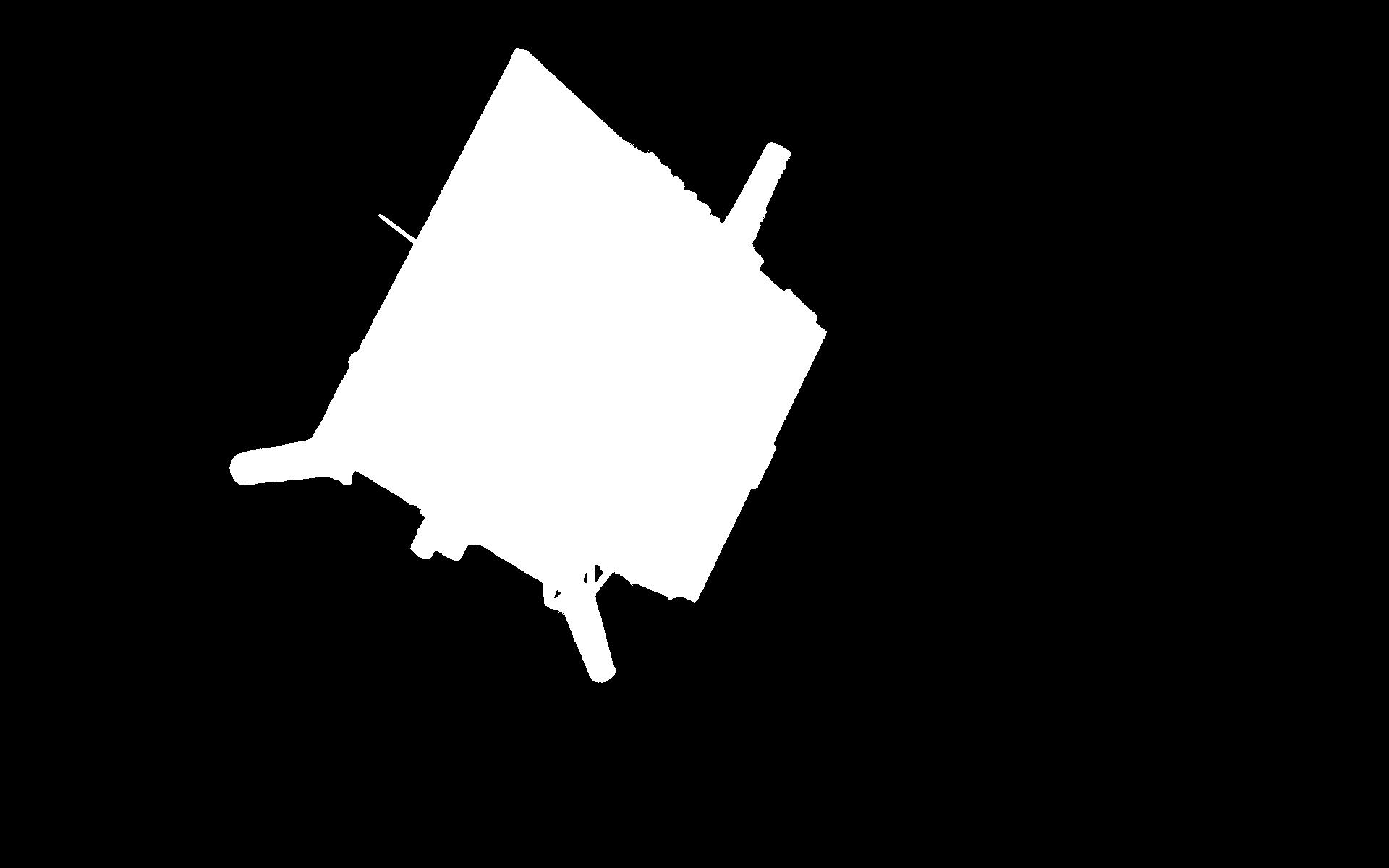}};
\node[right=0.6cm of pose.east] (output_pose) {\includegraphics[height=1.4cm]{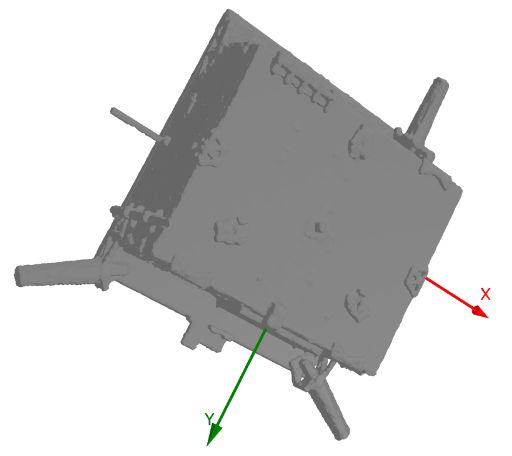}};

\node[anchor=west] at ($(output_pose.east) + (0.0cm, 0.0cm)$) {$(\hat{q},\hat{t})^{\text{reg}}$};
\node[anchor=west] at ($(output_heat.east) + (-0.075cm, -0.55cm)$) { $\mathbf{\hat{X}},\mathbf{\hat{Y}}$};
\node[anchor=west] at ($(output_segm.east) + (0.0cm, -0.25cm)$){ $\hat{m}$};

\draw[->] (segm.east) -- (output_segm.west);
\draw[->] (heatmap.east) -- (output_heat.west);
\draw[->] (pose.east) -- (output_pose.west);

\node[draw, rectangle, rounded corners, right=0.3cm of output_heat.east] (pnp) {\small PnP};
\node[right=0.3cm of pnp.east] (output_pose_PnP) {\includegraphics[height=1.4cm]{shirt/Pose_no_back_255.png}};
\node[anchor=north] at ($(output_pose_PnP.south) + (0.2cm, 0.1cm)$) { $(\hat{q},\hat{t})^{\text{heat}}$};

\draw[->] (output_heat.east) -- (pnp.west);
\draw[->] (pnp.east) -- (output_pose_PnP.west);

\end{tikzpicture}